\documentclass{article}

\pdfoutput=1

\usepackage{arxiv}

\usepackage{amsmath,amsfonts,bm}

\def\eqref#1{equation~\ref{#1}}

\def\floor#1{\lfloor #1 \rfloor}
\def\1{\bm{1}}

\def\mM{{\bm{M}}}

\DeclareMathAlphabet{\mathsfit}{\encodingdefault}{\sfdefault}{m}{sl}
\SetMathAlphabet{\mathsfit}{bold}{\encodingdefault}{\sfdefault}{bx}{n}
\newcommand{\tens}[1]{\bm{\mathsfit{#1}}}
\def\tA{{\tens{A}}}

\def\tE{{\tens{E}}}

\def\tG{{\tens{G}}}

\def\tM{{\tens{M}}}

\def\tP{{\tens{P}}}

\def\gG{{\mathcal{G}}}

\def\sV{{\mathbb{V}}}

\newcommand{\R}{\mathbb{R}}

\usepackage[T1]{fontenc}    
\usepackage{hyperref}       
\usepackage{url}            
\usepackage{booktabs}       
\usepackage{amsfonts}       
\usepackage{nicefrac}       
\usepackage{cleveref}       
\usepackage{graphicx}
\usepackage{natbib}
\usepackage{doi}
\usepackage{caption}
\usepackage{subcaption}
\usepackage{changepage}
\usepackage{relsize}

\graphicspath{{pldr_llm_figures/}}

\title{PLDR-LLM: Large Language Model from Power Law Decoder Representations}

\date{October 22, 2024}

\newif\ifuniqueAffiliation
\uniqueAffiliationtrue

\ifuniqueAffiliation 
\author{ \hspace{1mm}Burc Gokden \\
	Fromthesky Research Labs LLC\\
	Oregon, USA \\
	\texttt{burc@fromtheskyresearchlabs.com} \\
}
\else
\fi

\renewcommand{\headeright}{preprint}
\renewcommand{\undertitle}{}
\renewcommand{\shorttitle}{PLDR-LLM: Large Language Model from Power Law Decoder Representations}

\begin{document}
\maketitle

\begin{abstract}
We present the Large Language Model from Power Law Decoder Representations (PLDR-LLM), a language model that leverages non-linear and linear transformations through Power Law Graph Attention mechanism to generate well-defined deductive and inductive outputs. We pretrain the PLDR-LLMs of varying layer sizes with a small batch size of 32 and $\sim$8B tokens from the RefinedWeb dataset, and show that they achieve competitive performance in zero-shot and few-shot settings compared to scaled dot-product LLMs of similar model size reported in the literature. We show that deductive outputs of PLDR-LLMs can be used to compare model characteristics or improve the performance by introducing the Directed Acyclic Graph (DAG) loss as a metric and regularizer. Our results indicate that the initial maximum learning rate and warm-up steps have a lasting impact on deductive outputs throughout the pretraining. We provide a detailed description of PLDR-LLM architecture, its implementation and the pretraining procedure.
\end{abstract}

\section{Introduction}

Generative Pretrained Large Language Models (LLMs) demonstrated breakthrough improvements in performance on a wide range of Natural Language Processing (NLP) tasks such as common-sense reasoning, question answering, reading comprehension, and inference \citep{Radford2018gpt1, Radford2019gpt2, Brown2020gpt3, Touvron2023llama, Touvron2023llama2}. The performance gains in LLMs are typically achieved by increasing model size and  number of tokens for pretraining. Scaling laws for LLMs are extensively studied to optimize model size, number of tokens and the amount of compute required \citep{Kaplan2020scaling, Hoffmann2022chinchilla}. The quality of the training dataset also plays an important role in improving the model performance. Filtered and deduplicated web data \citep{Penedo2023falcon} and curated high quality data from books and technical papers \citep{Touvron2023llama2, Gao2020pile} considerably improves the model performance. For LLMs with sub-billion parameter range, performance was improved with deeper models and weight sharing strategies, as smaller models have a large share of parameter size accounted for input and output embedding layers \citep{Liu2024mobilellm}.

Current state-of-the-art LLMs rely on a model architecture that has deep layers of transformer decoders \citep{Vasvani2017} trained on large amount of text data in an unsupervised fashion. The transformer is a transductive model that utilizes scaled dot-product (SDP) attention for its encoder-decoder implementation to capture long range relationships within a context and avoid a bottleneck condition observed in recurrent neural networks. SDP attention leverages the linear transformations and the LLMs formed on SDP transformer decoder representations are also transductive in nature.

Power Law Graph Attention (PLGA) \citep{Gokden2021, Gokden2019} can be used to extend LLM architectures to have well-defined deductive and inductive outputs and utilize non-linear transformations besides linear ones in attending the input. They were first utilized in Power Law Graph Transformer architecture to demonstrate competitive performance for machine translation tasks \citep{Gokden2021}. The deductive outputs provide characteristic model parameters to observe the model response during inference as well as to regularize during pretraining to improve performance other than scaling the model size and number of tokens for pretraining. Regularizing deductive outputs also presents a means to modify model behavior as compared to solely scaling the size and data for performance as they can directly affect the model weights. 

For deductive outputs to be interpretable, they need to have a meaningful connection to input samples and retain this connection as much as possible through the design of the model architecture. This is achieved by treating the input sentence as a weighted graph $ \gG=\left( \sV , E \right) $ where tokens are nodes and N-dimensional embeddings form a feature vector for each node. The power law graph attention utilizes both non-linear and linear transformation steps to learn characteristics of the feature space as part of model parameters. The attention first learns a metric tensor for the manifold through a deep residual neural network. This metric tensor $\tA_{LM}$ is a representation for the structure of the N-dimensional manifold. The metric tensor can also be interpreted as a weighted adjacency matrix where embedding space dimensions are treated as vertices.  Power law coefficients $\tP$ are then learned to define strength and range of the interaction between each embedding dimension. Finally, the interactions of each dimension with all other dimensions in the manifold are learned as the energy-curvature tensor $\tG_{LM}$, which is a superposition of potentials. The final attention $\tE_{LM}$ is calculated by projecting the query and key vectors on $\tG_{LM}$. Attention $\tE_{LM}$ acts on the input as a linear transformation to generate an output in the form of graph $\gG$. The metric tensor, power coefficients (and the potential tensor, $\tA_{\textbf{P}}=\tA_{LM}^{\odot\tP}$) and energy-curvature tensor form part of the deductive outputs of power law graph attention we will focus in this implementation. This treatment enables us to approach the deductive outputs as representations of graphs that shape the manifold. 

In this paper, we present LLM architectures with deductive-inductive outputs that are designed by using the Power Law Decoder Representations (PLDR) which are based on the decoder layer of Power Law Graph Transformer. We make the following contributions:

\begin{itemize}

\item A new deductive-inductive large language model architecture, the Large Language Model from Power Law Decoder Representations (PLDR-LLM) is proposed and implemented.

\item PLDR-LLMs are shown to have competitive results compared to reference scaled dot-product LLMs of similar model size that are pretrained with larger amount of tokens, larger batch size and longer context length in the literature. We demonstrate that model is highly robust to gradient noise and a global batch size of 32 (16 batches per rank) and $\sim$8B tokens are enough to pretrain competitive LLMs with a context length of 1024 tokens.

\item We study Directed Acyclic Graph (DAG) loss \citep{Zheng2018notears} of deductive outputs as a metric and regularizer to observe PLDR-LLM characteristics during training and show that deductive outputs can be regularized to approximate a DAG and improve benchmark scores over an unregularized base model.

\item We introduce a semi-positive activation function we refer to as iSwiGLU which is SwiGLU with identity matrix as weights and no bias.

\item Implementation of PLDR-LLM architecture and framework is available at:\\ \url{https://github.com/burcgokden/LLM-from-Power-Law-Decoder-Representations}

\end{itemize}

\section{Approach}

Our training approach is similar to other generative LLMs that implement attention based decoders \citep{Radford2019gpt2, Touvron2023llama, Touvron2023llama2}. We trained the PLDR-LLMs autoregressively in an unsupervised manner while minimizing the cross-entropy loss. In addition to evaluating zero-shot and few-shot benchmarks, we did ablation studies on the DAG loss of deductive outputs and applied DAG regularization on PLDR-LLMs. 

\subsection{Model Architecture}

The model architecture is based on Power Law Graph Transformer with following parameters and modifications to attention implementation: 
\begin{itemize}

\item The $d_{k}=d_{model}/h$ parameter is set at $64$ for each model, for embedding dimension $d_{model}$ and number of attention heads $h$. 

\item The ReLU activation in the residual neural network for learning metric tensor $\tA_{LM}$ and feedforward network is replaced with SwiGLU activation function \citep{Shazeer2020swiglu, Dauphin2017glu}. For metric learner we use 8 layers of residual units with 2 SwiGLU feedforward networks (FFNs) in each unit. The layer sizes for gated-linear and linear-only layers of a single SwiGLU network were 170:64 ($\floor{\frac{2}{3} 4d_{k}}$:$d_{k}$) and 300:112 for PLDRv5 and PLDRv9 flavors of the model, respectively. For the feedforward network at the end of each attention layer, the gated-linear layer size was set as $\frac{2}{3} 4d_{model}$ following the same scaling approach in \citep{Shazeer2020swiglu}. 

\item The LeakyRELU activation applied after query and key projections onto $\tG_{LM}$ is removed from the model. 

\item The ReLU activation at the end of metric learner was replaced with iSwiGLU activation function: $x\odot \sigma(x)$ where $\sigma()$ is the Swish (SiLU) activation function \citep{Ramachandran2017swish, Elfwing2018silu}. 

\item The positional embeddings are replaced with rotary embeddings \citep{Su2021roformer} at each layer of attention mechanism. 

\item LayerNorm \citep{Ba2016layer} layers were added after the embedding layer and factoring of its output by $\sqrt{d_{model}}$ and after the linear self-attention of query vectors that is input to the residual network. 

\item Drop-out was not used at any layer.

\end{itemize}

A diagram of PLDR-LLM architecture is shown in the appendix. Model hyperparameters are shown in Table \ref{table1}.

The learning rate was set at $1\times10^{-3}$ for all models except for PLDRv5-DAG-1 and PLDRv5-DAG-3 which had a learning rate of $1.2\times10^{-3}$; and PLDRv5-ab-1 which had a learning rate of $6\times10^{-4}$ for ablation. The warm-up step size was set at 2000 steps for all models except for PLDRv5-1, PLDRv9-1 and PLDRv5-tab-2 which had a warm-up step size of 8000. The models were trained using AdamW optimizer \citep{Loshchilov2019adamw} with parameters $\beta_{1}=0.9, \, \beta_{2}=0.95, \, eps=1\times10^{-5}$ and gradient clipping of $1.0$ and weight decay set at $0.1$ \citep{Touvron2023llama, Touvron2023llama2}. The learning rate decay was set to follow cosine annealing schedule with linear warm-up \citep{Loshchilov2016sgdr}. The final learning rate was set at $10\%$ of maximum learning rate \citep{Hoffmann2022chinchilla}. All models were trained on two RTX $4090$ GPUs with $24$ GB of RAM. Batch size and context length was set at $32$ and $1024$ tokens.

All PLDR-LLMs trained are memory constrained such that they are designed to fit in a single GPU memory. As we increase the number of decoder layers, the number of attention heads per layer and parameter size of models are reduced. Since deep layers of neural networks are often associated with better generalization, we are also interested in training PLDR-LLMs with highest number of decoder layers possible. The model with highest number of decoder layers, PLDRv5-1, utilizes the GPU memory extensively and the reduced number of heads per layer makes this model more susceptible to instabilities. As a result, we were able to achieve stable training for this model at a longer warm-up step size.

\begin{table}[ht]
\caption{Set of hyperparameters for models pretrained. "DAG" labeled models are regularized with DAG loss of deductive outputs. "ab" labeled model is for ablation of small learning rate, and "tab" labeled models are for ablation of warm-up steps with a different tokenizer model.}
\label{table1}
\centering
\begin{tabular}{c c c c c c c c c}
\toprule[1.2pt] 
Model &	 \# Parameters & \# Layers  & \# Heads & $d_{model}$ & SwiGLU:Linear & \multicolumn{1}{p{1.5cm}}{\centering Learning \\ Rate} & \multicolumn{1}{p{1.5cm}}{\centering Warm-Up \\ Steps} \\ 
\midrule[1.2pt]
PLDRv5-1 & 104M	 & 7 & 12 & 768	 & 170:64 &  $1\times10^{-3}$ & 8000 \\
\midrule
PLDRv5-2 & 110M & 5 & 14 & 896 & 170:64 &  $1\times10^{-3}$ & 2000 \\ 
\midrule 
PLDRv5-3 & 144M	 & 3 & 20 & 1408 & 170:64 & $1\times10^{-3}$ & 2000 \\ 
\midrule 
PLDRv5-4 & 260M & 1 & 42 & 2688 & 170:64 &  $1\times10^{-3}$ & 2000 \\
\midrule[1.2pt]
PLDRv9-1 & 114M & 4 & 15 & 960 & 300:112 & $1\times10^{-3}$ & 8000 \\   
\midrule 
PLDRv9-2 & 147M & 3 & 20 & 1280 & 300:112 & $1\times10^{-3}$ & 2000 \\  
\midrule[1.2pt] 
PLDRv5-DAG-1 & 110M & 5 & 14 & 896 & 170:64 & $1.2\times10^{-3}$ & 2000 \\ 
\midrule 
PLDRv5-DAG-2 & 110M & 5 & 14 & 896 & 170:64 & $1\times10^{-3}$ & 2000 \\ 
\midrule
PLDRv5-DAG-3 & 110M & 5 & 14 & 896 & 170:64 & $1.2\times10^{-3}$ & 2000 \\ 
\midrule 
PLDRv5-DAG-4 & 110M & 5 & 14 & 896 & 170:64 & $1\times10^{-3}$ & 2000 \\
\midrule 
PLDRv5-DAG-5 & 110M & 5 & 14 & 896 & 170:64 & $1\times10^{-3}$ & 2000 \\ 
\midrule 
PLDRv5-DAG-6 & 110M & 5 & 14 & 896 & 170:64 & $1\times10^{-3}$ & 2000  \\
\midrule 
PLDRv5-DAG-7 & 110M & 5 & 14 & 896 & 170:64 & $1\times10^{-3}$ & 2000 \\
\midrule[1.2pt]
PLDRv5-ab-1 & 110M & 5 & 14 & 896 & 170:64 & $6\times10^{-4}$ & 2000 \\
\midrule
PLDRv5-tab-1 & 110M & 5 & 14 & 896 & 170:64 & $1\times10^{-3}$ & 2000 \\
\midrule
PLDRv5-tab-2 & 110M & 5 & 14 & 896 & 170:64 & $1\times10^{-3}$ & 8000 \\
\bottomrule[1.2pt]
\end{tabular}
\end{table}

\subsection{Directed Acyclic Graph (DAG) Regularization}

We add mean absolute value of DAG loss of deductive outputs $\tA_{LM}$, $\tA_{\textbf{P}}$, $\tG_{LM}$ from all attention heads to the cross-entropy loss for pretraining of several PLDR-LLMs. This loss condition aims to reduce the number of cyclic paths of any length to zero for deductive outputs and acts as a regularizer. We also observe DAG loss as a metric to compare models without regularization. DAG loss was first introduced in \citep{Zheng2018notears} for causal modeling of graphs with a smooth and differentiable loss function.

A square matrix $\mM \in \R^{d \times d}$ is a Directed Acyclic Graph if it satisfies the condition:
\begin{equation}
\label{eq1}
     tr(e^{\mM \odot \mM})-d=0
\end{equation}

The the DAG regularizer $DLR(\tA_{LM}, \tA_{P}, \tG_{LM})$ added to the PLDR-LLM cross-entropy loss is defined as:
\begin{align}
\label{eq2}
DL(\tM) &= \frac{1}{BLh}\sum_{B, L, h} \left\lvert \log{\frac{tr\left( e^{\tM \odot \tM} \right)}{d_{k}}} \right\rvert \\
DLR(\tA_{LM}, \tA_{P}, \tG_{LM}) &= \lambda_{DAG1}DL(\tA_{LM})+\lambda_{DAG2}DL(\tA_{P})+\lambda_{DAG3}DL(\tG_{LM})
\end{align}

where $\tM$ is a tensor and $B$, $L$ are the batch size and number of decoder layers. $\lambda_{DAG1}$ , $\lambda_{DAG2}$ and $\lambda_{DAG3}$ are coefficients that determine the strength of regularization.

\section{Dataset}

We used the first $\sim$8B tokens from the RefinedWeb dataset for pre-training. RefinedWeb is a publicly available, high quality web based English dataset with extensive deduplication and filtering \citep{Penedo2023falcon}. The tokenizer we used was SentencePiece unigram tokenizer \citep{Kudo2018sentencepiece, Kudo2018subword} with vocabulary size of $32000$. We trained the tokenizer model by randomly sampling from a portion of RefinedWeb dataset. The tokenizer was set to split all digits into single units and to fallback to bytes to decompose unknown UTF-8 characters following similar approach in \citep{Touvron2023llama}. The tokenizer allows padding and only an "[END]" token was added at the end of sentence during tokenization. 

Each sample in the dataset was first tokenized and formed into larger batches which were then concatenated. The concatenated samples were chunked into contiguous sets of $1024$ tokens and batched with final batch size of $16$ per rank. Occasionally appearing chunks with tokens less than $1024$ are padded with "[PAD]" token. The model implementation was designed to ignore padding during pretraining and evaluation of metric values.

\section{Experiments}
We evaluated PLDR-LLMs with varying size of model parameters from 104M to 260M by changing the layer size, number of heads and residual unit SwiGLU FFN layer sizes (Table \ref{table1}). Each PLDR-LLM was pretrained with first $\sim$8B tokens of RefinedWeb dataset in the same order and on a single epoch. For ablation studies, we used the PLDRv5-2 as a base model. We pretrained the base model with low learning rate, longer warm-up steps and a different tokenizer model trained with same parameters on the RefinedWeb dataset. The DAG regularization was applied on a 5-layer, 14-head PLDR-LLM on the deductive inputs and compared to a base model (PLDRv5-2) with same hyperparameters without any regularization. DAG regularization strength was skewed from 0.001 to 1 for $\lambda_{DAG1}$ and from 0.005 to 0.05 for $\lambda_{DAG2}$ and $\lambda_{DAG3}$ on PLDRv5-DAG models (Table \ref{table2}).

We observed the training loss/accuracy and DAG loss of deductive outputs and evaluated the PLDR-LLMs on a range of tasks for commonsense reasoning, question answering and language understanding. The models were evaluated with tinyBenchmarks version of datasets \citep{Polo2024tinybenchmarks} available in Eleuther AI Evaluation Harness Suite \citep{evalharness} for zero-shot and few-shot performance. The PLDR-LLM implementation used in our experiments is not optimized for fast inference, and tinyBenchmarks datasets provide a quick way to evaluate model for few-shot performance that is also compatible with the context length of the pretrained PLDR-LLMs. We also evaluated our LLMs with a set of full-length benchmark datasets with only zero-shot setting. For comparison, several LLMs of similar size reported in literature (Cerebras-GPT-111M\footnote{https://huggingface.co/cerebras/Cerebras-GPT-111M} \citep{dey2023cerebras}, GPT2-124M\footnote{https://huggingface.co/openai-community/gpt2} \citep{Radford2019gpt2}, GPT-Neo-125M\footnote{https://huggingface.co/EleutherAI/gpt-neo-125m} \citep{gptneo,Gao2020pile} and Phythia-160M\footnote{https://huggingface.co/EleutherAI/pythia-160m-deduped} \citep{Biderman2023pythia}) were evaluated with same benchmark configurations using their implementations available on Huggingface platform. The evaluation metric for the benchmarks except for TruthfulQA were byte-length normalized accuracy to ensure the reported results are tokenization agnostic. For TruthfulQA \citep{Lin2021truthfulqa}, a normalized metric for multiple choice, multiple true answers is used. We report on the average accuracies without (Avg-1) and with (Avg-2) TruthfulQA included since this benchmark usually tends to perform worse for models that learn the training distribution better such as larger models.

\section{Results}

\begin{figure}
\centering
\begin{subfigure}[b]{0.45\textwidth}
	\centering
	\includegraphics[width=1\textwidth]{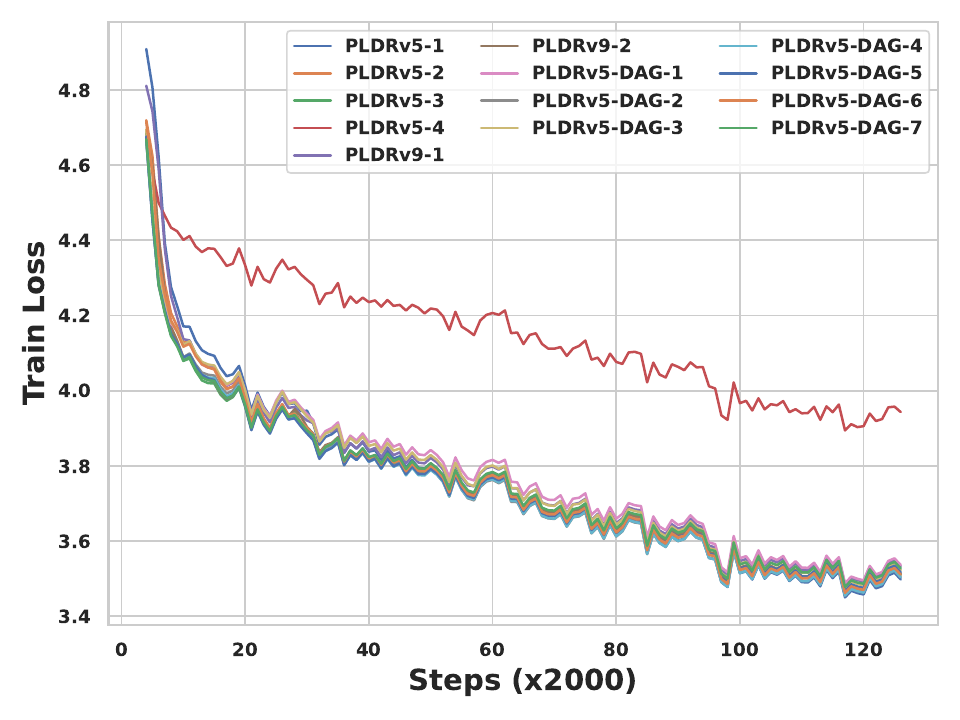}
	\caption{}
	\label{fig1a}
\end{subfigure}
\hfill
\begin{subfigure}[b]{0.45\textwidth}
	\centering
	\includegraphics[width=1\textwidth]{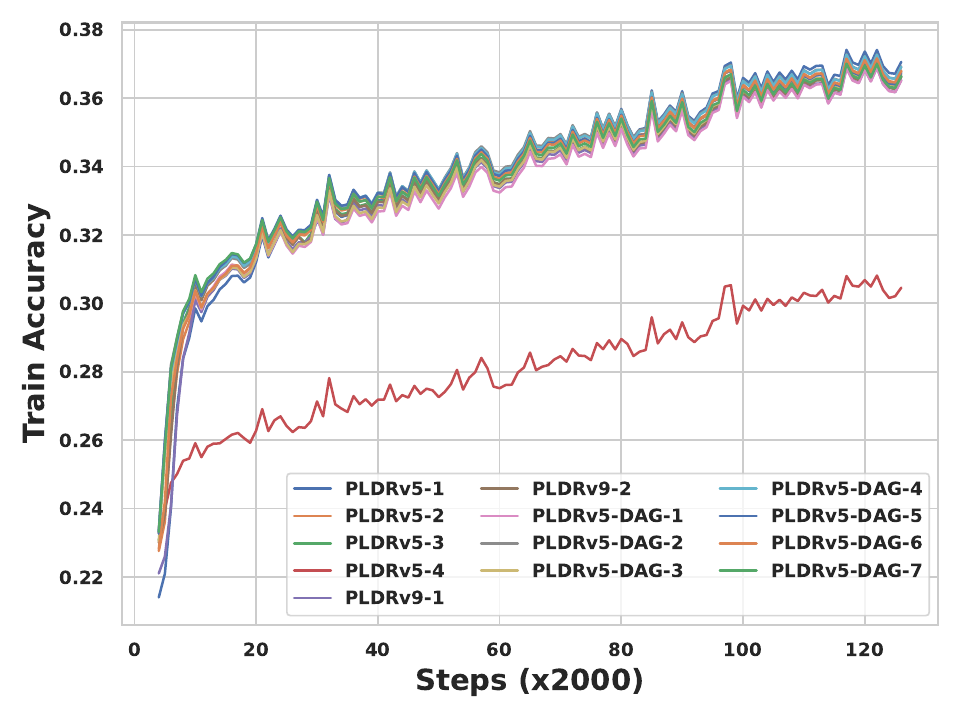}
	\caption{}
	\label{fig1b}
\end{subfigure}
\hfill
\begin{subfigure}[b]{0.45\textwidth}
	\centering
	\includegraphics[width=1\textwidth]{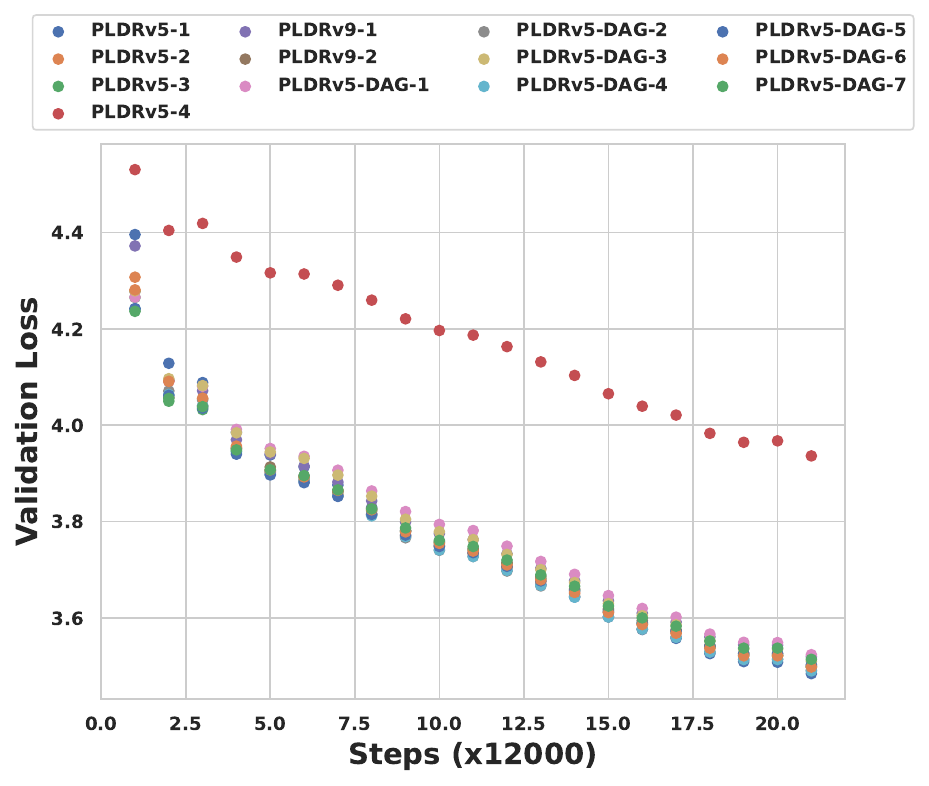}
	\caption{}
	\label{fig1c}
\end{subfigure}
\hfill
\begin{subfigure}[b]{0.45\textwidth}
	\centering
	\includegraphics[width=1\textwidth]{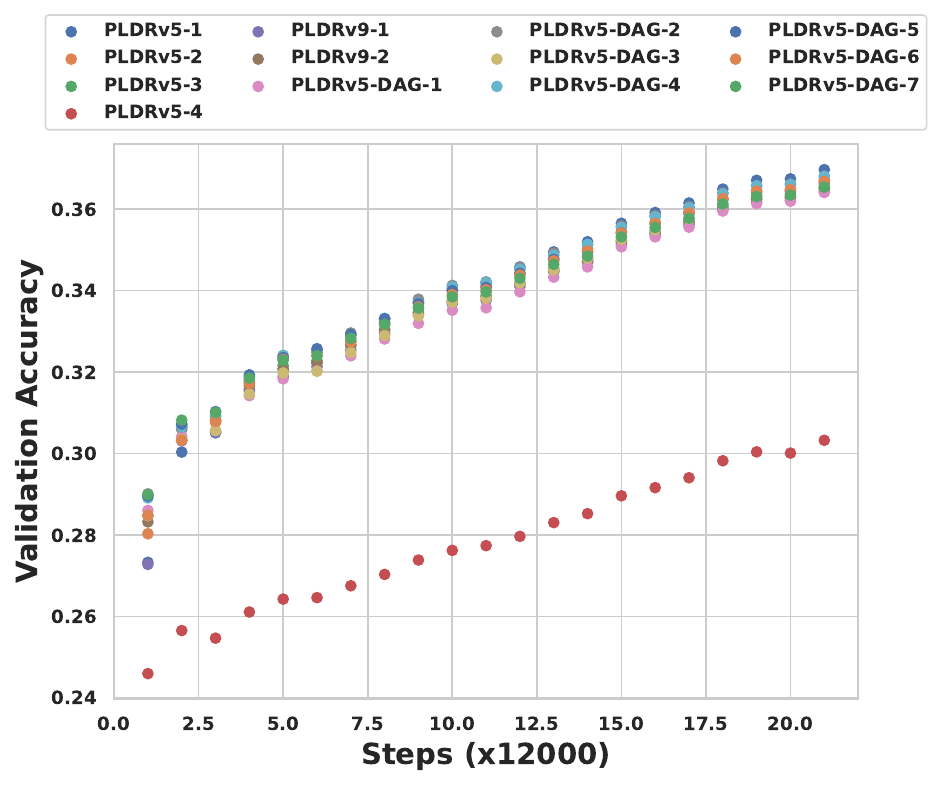}
	\caption{}
	\label{fig1d}
\end{subfigure}
\caption{Train and validation loss/accuracy curves for PLDR-LLMs. Train loss is captured as a running loss at every 2000 steps. Validation loss is measured at every 12000 steps using 2000 batches/rank from part of RefinedWeb dataset that is not used in pretraining.}
\label{fig1}
\end{figure}

\textbf{Loss/Accuracy Curves.} The loss and accuracy curves are shown in Fig. \ref{fig1} for training and validation. For a small batch size of 32, the loss and accuracy curves were quite robust and the models are underfit. We observed that running training loss values and validation losses are in good agreement for both regularized and unregularized PLDR-LLMs. All models except for the single decoder layer model (PLDRv5-4) converged to a loss value around $3.5$ after 250k training steps for the RefinedWeb dataset and hyperparameters we used in the experiments. PLDRv5-4 exhibits a similar trend as deeper models but it converges to a higher loss value around $3.9$.

\begin{figure}
\centering
\begin{subfigure}[b]{0.45\textwidth}
	\centering
	\includegraphics[width=1\textwidth]{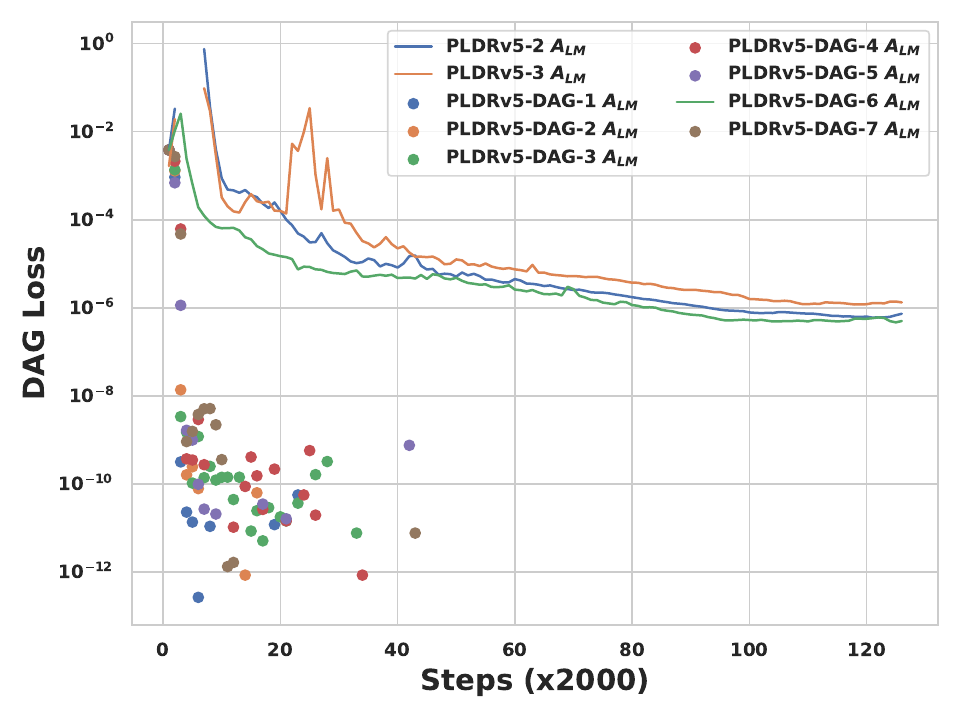}
	\caption{}
	\label{fig2a}
\end{subfigure}
\begin{subfigure}[b]{0.45\textwidth}
	\centering
	\includegraphics[width=1\textwidth]{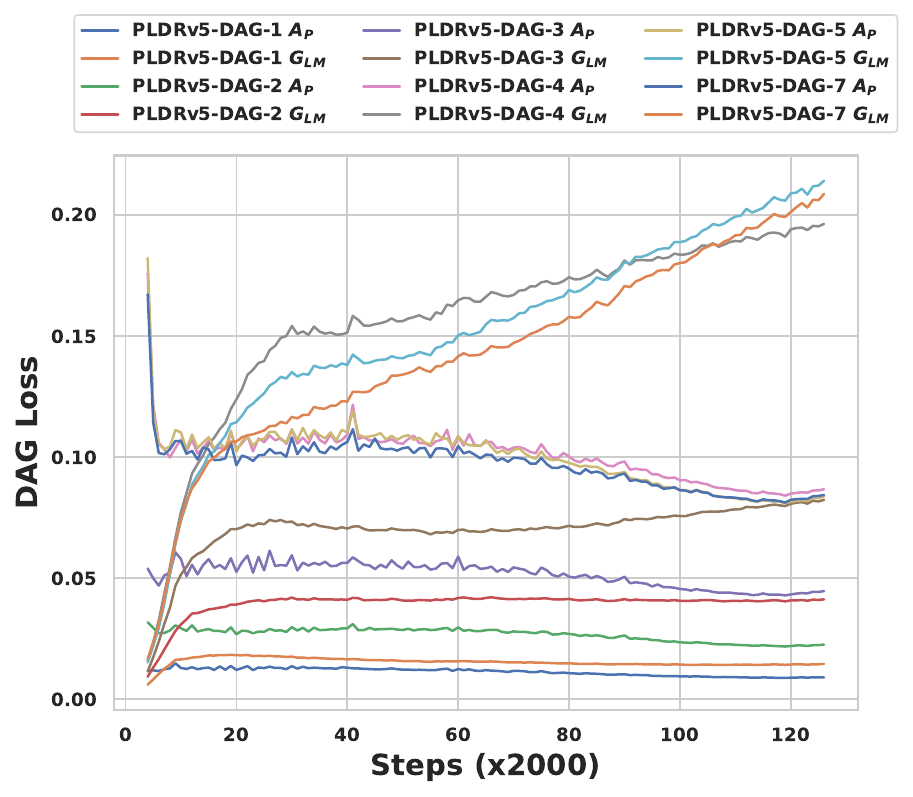}
    \caption{}
	\label{fig2b}
\end{subfigure}
\hfill
\caption{DAG regularizer log loss trend during pretraining before scaling with regularizer coefficients: (a) for $\tA_{LM}$, (b) for $\tA_{\textbf{P}}$ and $\tG_{LM}$. For unregularized models, DAG loss for $\tA_{LM}$ overflows for a few thousand steps after warm-up. For regularized models, DAG loss for $\tA_{LM}$ goes to zero quickly. Overflow and zero values are omitted on the log scale axis for the loss. } 
\label{fig2}
\end{figure}

\noindent\textbf{DAG Loss on Deductive Outputs.} For the tokenizer model used for pretraining, the metric tensor $\tA_{LM}$ typically follows a low DAG loss with little divergence for the base model without regularization, whereas the potential, and energy-curvature tensors have a DAG loss that diverges beyond the floating point precision (overflow). This observation holds for other PLDR-LLMs at inference time (Table \ref{table2}). We apply the DAG loss as a regularizer to condition other deductive outputs and observe the effect of tensors that approximate a DAG on the model performance of benchmarks. When regularized, we see that the DAG loss recovers from the overflow condition for deductive outputs, and makes it possible to pretrain models over a range of regularization coefficients\footnote{If used as a metric, scaling the deductive outputs before evaluating DAG loss may help recover from overflow, though we chose not to explore this approach here.}. The sum of contributions of potential and energy-curvature tensor DAG losses scaled by the regularization coefficients converge to loss values around $1-1.5\times10^{-3}$.

\begin{table}[ht]
\caption{DAG regularization coefficients and DAG loss values for deductive outputs observed at inference by generating 50 tokens at $\text{top-k}=1$ (greedy sampling) for the input: "Write a letter requesting that people use language models responsibly." NA means regularization is not applied for respective deductive output. $\nearrow$ indicates that the DAG loss diverges and results in overflow. }
\label{table2}
\centering
\begin{tabular}{c c c c c c c c}
\toprule[1.2pt] 
Model  &$\lambda_{DAG1}$ & $\lambda_{DAG2}$ & $\lambda_{DAG3}$ & \multicolumn{3}{c}{DAG Loss at Inference}  & $DLR(\tA_{P}, \tG_{LM})$\\
\cmidrule(lr){5-7}
& & & & DL($\tA_{LM})$ & DL($\tA_{P})$ & DL($\tG_{LM})$ \\
\midrule[1.2pt]
PLDRv5-1 & NA & NA & NA & $6.77\times10^{-7}$ & $\nearrow$ & $\nearrow$ & NA \\ 
\midrule 
PLDRv5-2 & NA & NA & NA & $7.51\times10^{-7}$ & $\nearrow$ & $\nearrow$ & NA \\ 
\midrule 
PLDRv5-3 & NA & NA & NA & $8.01\times10^{-5}$ & $\nearrow$ & $\nearrow$ & NA \\
\midrule 
PLDRv5-4 & NA & NA & NA & $1.78\times10^{-6}$ & $\nearrow$ & $\nearrow$  & NA \\
\midrule[1.2pt]  
PLDRv9-1 & NA & NA & NA & $2.03\times10^{-6}$ & $\nearrow$ & $\nearrow$ & NA \\
\midrule 
PLDRv9-2 & NA & NA & NA & $1.47\times10^{-6}$ & $\nearrow$ & $\nearrow$ & NA \\ 
\midrule[1.2pt] 
PLDRv5-DAG-1 & 0.05 & 0.05 & 0.05 & 0 & $8.61\times10^{-3}$ & $1.43\times10^{-2}$ & $1.15\times10^{-3}$ \\ 
\midrule 
PLDRv5-DAG-2 &  0.02 & 0.02 & 0.02 & 0 & $2.22\times10^{-2}$ & $4.03\times10^{-2}$ & $1.25\times10^{-3}$  \\ 
\midrule
PLDRv5-DAG-3 &0.01 & 0.01 & 0.01 & 0 & $4.36\times10^{-2}$ & $8.03\times10^{-2}$ & $1.24\times10^{-3}$ \\ 
\midrule 
PLDRv5-DAG-4 &0.005 & 0.005 & 0.005 & 0 & $8.51\times10^{-2}$ & $1.94\times10^{-1}$ & $1.39\times10^{-3}$ \\ 
\midrule 
PLDRv5-DAG-5 & 1 & 0.005 & 0.005 & 0 & $8.19\times10^{-2}$ & $2.15\times10^{-1}$ & $1.48\times10^{-3}$ \\ 
\midrule 
PLDRv5-DAG-6 & 0.005 & NA & NA & $5.09\times10^{-7}$ & $\nearrow$ & $\nearrow$ & NA \\ 
\midrule 
PLDRv5-DAG-7 & 0.001 & 0.005 & 0.005 & 0 & $8.31\times10^{-2}$ & $2.07\times10^{-1}$ & $1.45\times10^{-3}$  \\  
\midrule[1.2pt]
PLDRv5-tab-1 & NA & NA & NA & $4.24\times10^{-3}$ & $\nearrow$ & $\nearrow$ & NA \\
\midrule 
PLDRv5-tab-2 & NA & NA & NA & $2.58\times10^{-3}$ & $\nearrow$ & $\nearrow$ & NA \\
\bottomrule[1.2pt]
\end{tabular}
\end{table}

The DAG regularizer loss on metric tensor converges to zero\footnote{Convergence to zero is also limited by the floating point precision (underflow).} (Fig. \ref{fig2a}) rather quickly when there is a non-zero regularizer on potential and energy-curvature tensors. When applied only on metric tensor, the DAG regularizer converges slowly to lower values similar to unregularized PLDR-LLMs and does not go down to zero early in pretraining. The contributions of regularizing losses over the course of pretraining are shown in Fig. \ref{fig2b} for potential and energy-curvature tensors. Compared to loss/accuracy curves which tend to agree tightly, the DAG losses of deductive outputs show clear distinctions for each model.

\begin{figure}
\centering
\begin{subfigure}[b]{0.45\textwidth}
	\centering
	\includegraphics[width=1\textwidth]{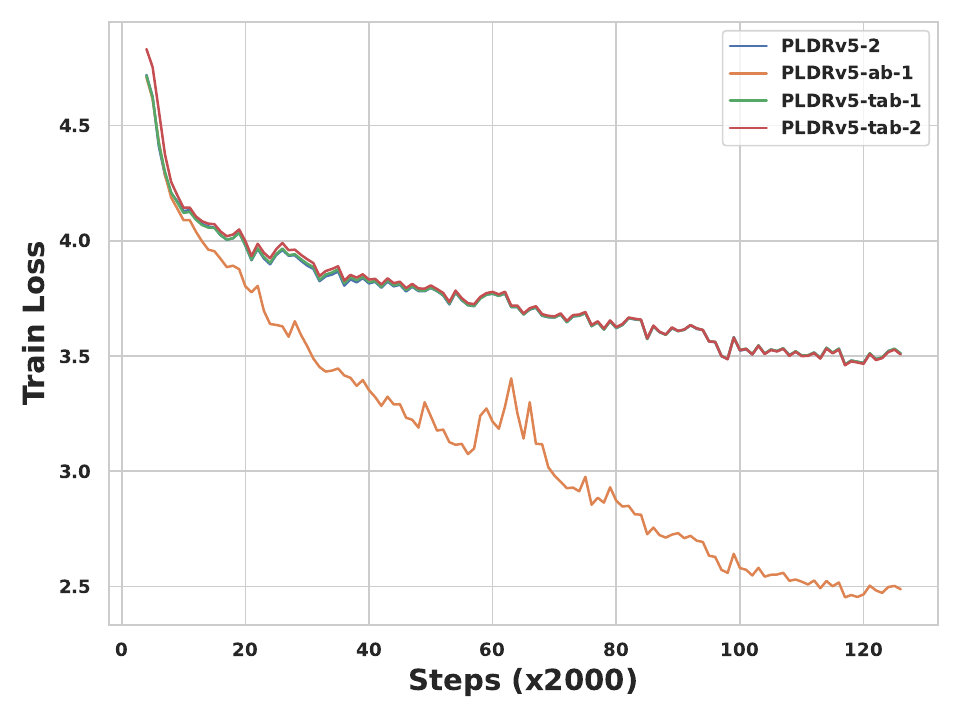}
	\caption{}
	\label{fig3a}
\end{subfigure}
\hfill
\begin{subfigure}[b]{0.45\textwidth}
	\centering
	\includegraphics[width=1\textwidth]{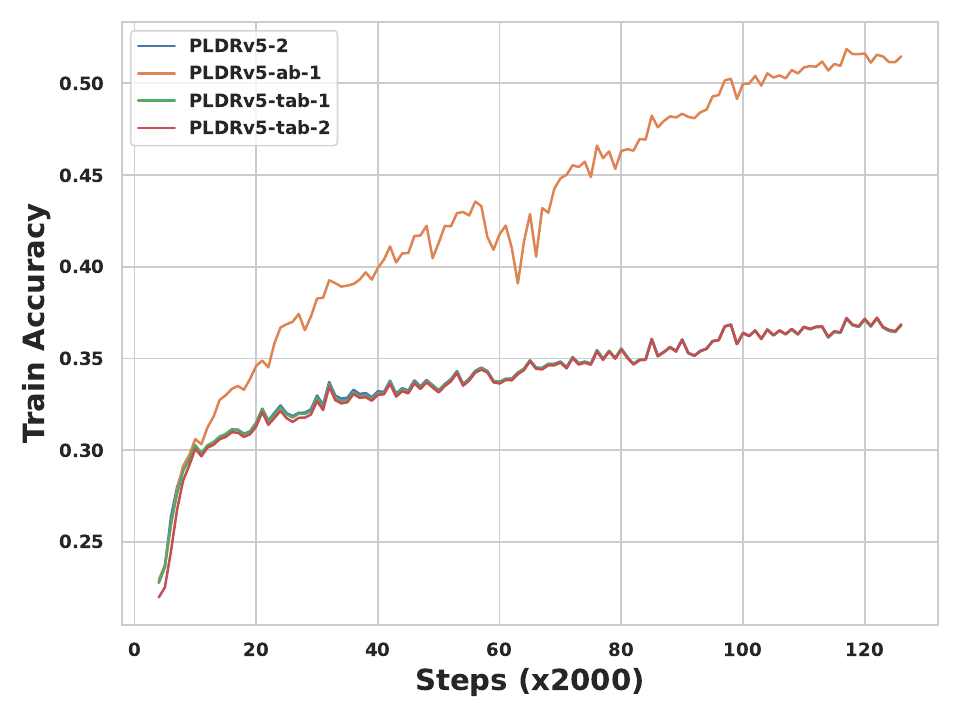}
	\caption{}
	\label{fig3b}
\end{subfigure}
\hfill
\begin{subfigure}[b]{0.45\textwidth}
	\centering
	\includegraphics[width=1\textwidth]{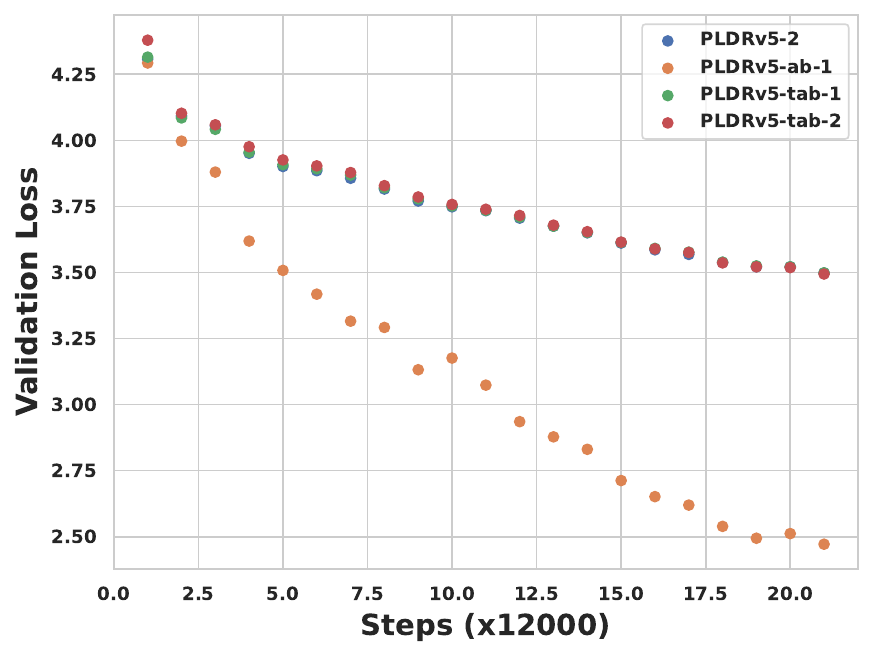}
	\caption{}
	\label{fig3c}
\end{subfigure}
\hfill
\begin{subfigure}[b]{0.45\textwidth}
	\centering
	\includegraphics[width=1\textwidth]{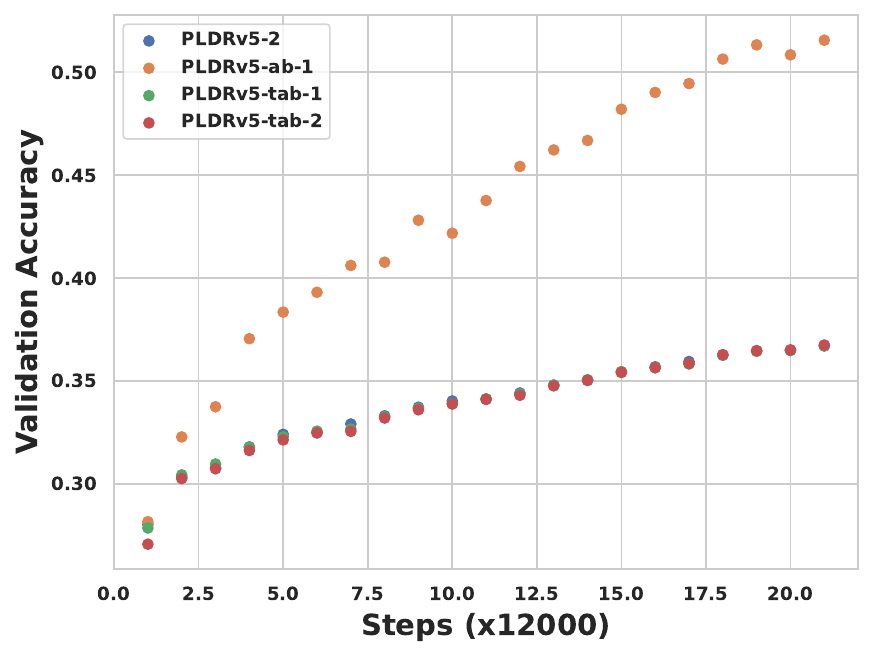}
	\caption{}
	\label{fig3d}
\end{subfigure}
\begin{subfigure}[b]{0.45\textwidth}
	\centering
	\includegraphics[width=1\textwidth]{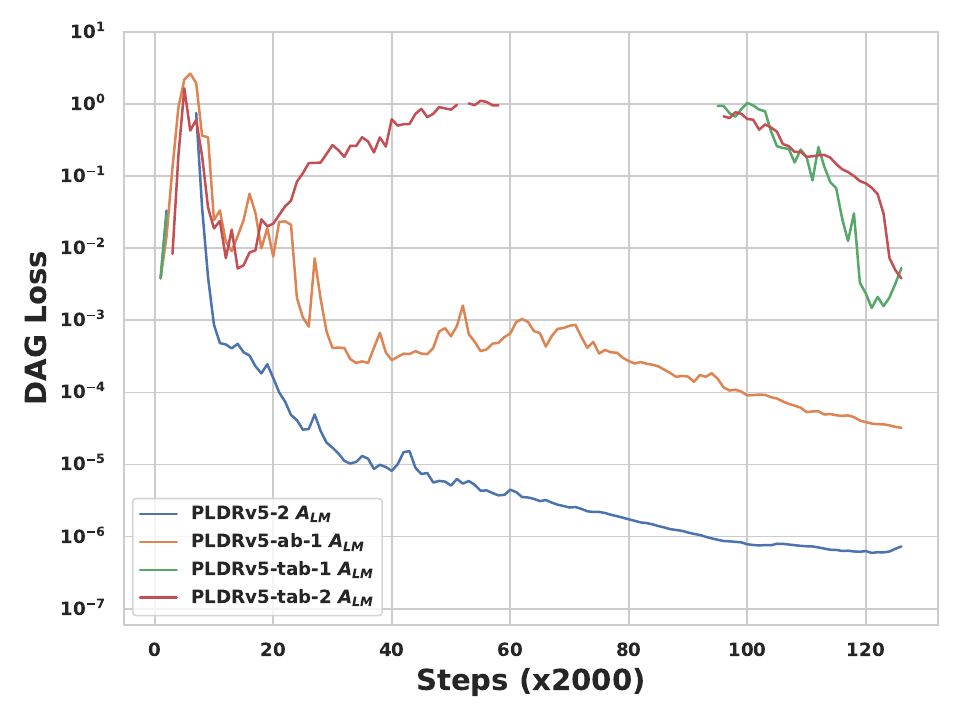}
	\caption{}
	\label{fig3e}
\end{subfigure}
\caption{(a)-(d) Train and validation loss/accuracy curves and (e) DAG loss for $\tA_{LM}$ for ablation of PLDR-LLMs for low learning rate, longer warm-up steps and different tokenizer model. The discontinuities in curves is due to overflow of DAG loss value.}
\label{fig3}
\end{figure}

\noindent\textbf{tinyBenchmarks Results.} We evaluated PLDR-LLMs and reference LLMs on tinyBenchmarks versions of ARC-c \citep{Clark2018arc}, Hellaswag \citep{Zellers2019hellaswag}, MMLU \citep{Hendrycks2021test, Hendrycks2021ethics}, Winogrande \citep{Keisuke2019winogrande} and TruthfulQA datasets. All datasets except for tinyMMLU and tinyTruthfulQA were evaluated on few-shot setting. The results are shown in Table \ref{table3}. 

The highest average score with and without tinyTruthfulQA is achieved by the deepest and thinnest model (PLDRv5-1) with the smallest model size at 104M, better than the reference models and DAG regularized models. We see an uptick in average scores for PLDRv5-3, which is the largest parameter size model with multiple decoder layers. The widest, single decoder layer PLDR-LLM (PLDRv5-4) with the largest model size at 260M has the lowest average scores. This result underscores that, with other hyperparameters such as dataset and batch size staying the same, scaling parameter size alone is not enough, smaller models with deeper and thinner layers can outperform larger models.

The PLDR-LLMs with DAG regularization achieve highest benchmark results for tinyMMLU (PLDRv5-DAG-4), tinyWinoGrande (PLDRv5-DAG-3) and tinyTruthfulQA (PLDRv5-DAG-6) compared to the base model (PLDRv5-2) and reference models. On average, DAG regularized models improve over the base model performance for several different regularization settings (PLDRv5-DAG-1 to 4, and 7) with PLDRv5-DAG-7 demonstrating the largest Avg-1 score over the base model. This model also has the lowest tinyTruthfulQA score among DAG regularized models. PLDRv5-DAG-3 has the largest Avg-2 score over the base model. In our evaluation, tinyTruthfulQA performance was higher for the reference model that has the smallest model size and lowest Avg-1 score and the DAG regularized model which has second lowest Avg-1 score among DAG regularized models. For tinyARC-c and  tinyHellaswag benchmarks the reference models have the highest scores and PLDR-LLMs with DAG regularization do not show any significant improvement over the base model.

\begin{table}[!htb]
\caption{Tinybenchmarks evaluation results. For the datasets that were evaluated in a few-shot setting, number of few-shots are shown within parantheses next to the dataset name. tARC-c: tinyARC-Challenge, tHS: tinyHellaswag, tMMLU: tinyMMLU, tWG: tinyWinoGrande, tTQA: tinyTruthfulQA.}
\label{table3}
\centering
\begin{tabular}{c c c c c c c c }
\toprule[1.2pt] 
Model & tARC-c(25) & tHS(10) & tMMLU & tWG(5) & Avg-1 & tTQA & Avg-2 \\ 
\midrule[1.2pt]
Cerebras-GPT-111M & 25.35 & 34.79 & 28.78 & 44.79 & 33.43 & 47.41 & 36.23 \\
\midrule 
GPT-2-124M & 27.43 & \textbf{38.79} & 29.71 & 42.70 & 34.66 & 41.56 & 36.04 \\  
\midrule 
GPT-Neo-125M & 31.34 & 37.78 & 26.33 & 47.23 & 35.67 & 43.19 & 37.18 \\
\midrule 
Pythia-160M & \textbf{31.73} & 30.93 & 30.09 & 47.83 & 35.15 & 44.09 & 36.93 \\ 
\midrule[1.2pt]  
PLDRv5-1 & 29.06 & 32.49 & 31.15 & 51.70 & \textbf{36.10} & 44.09 & \textbf{37.70} \\ 
\midrule 
PLDRv5-2 & 28.35 & 30.95 & 26.78 & 47.02 & 33.28 & 44.90 & 35.60 \\
\midrule 
PLDRv5-3 & 30.63 & 29.16 & 30.15 & 50.62 & 35.14 & 44.41 & 36.99 \\ 
\midrule  
PLDRv5-4 & 22.71 & 29.98 & 27.67 & 45.69 & 31.51 & 46.71 & 34.55 \\ 
\midrule[1.2pt]
PLDRv9-1 & 26.51 & 29.13 & 26.33 & 48.82 & 32.70 & 45.29 & 35.22 \\  
\midrule 
PLDRv9-2 & 26.36 & 29.05 & 30.48 & 51.25 & 34.28 & 44.93 & 36.41 \\ 
\midrule[1.2pt]
PLDRv5-DAG-1 & 29.47 & 29.54 & 28.24 & 47.82 & 33.77 & 45.54 & 36.12 \\ 
\midrule
PLDRv5-DAG-2 & 28.05 & 32.06 & 26.51 & 46.87 & 33.37 & 45.20 & 35.74 \\
\midrule
PLDRv5-DAG-3 & 26.74 & 29.05 & 29.00 & \textbf{52.72} & 34.38 & 44.74 & 36.45 \\
\midrule
PLDRv5-DAG-4 & 26.41 & 28.49 & \textbf{33.06} & 46.10 & 33.51 & 43.83 & 35.58 \\
\midrule
PLDRv5-DAG-5 & 26.38 & 31.85 & 28.16 & 42.96 & 32.34 & 45.56 & 34.98 \\
\midrule
PLDRv5-DAG-6 & 24.00 & 30.70 & 27.47 & 47.78 & 32.49 & \textbf{48.20} & 35.63 \\
\midrule
PLDRv5-DAG-7 & 28.42 & 31.41 & 30.64 & 48.48 & 34.74 & 42.77 & 36.34 \\
\midrule[1.2pt]
PLDRv5-tab-1 & 30.59 & 28.82 & 28.36 & 48.17 & 33.99  & 43.64 & 35.92 \\
\midrule
PLDRv5-tab-2 & 26.64 & 30.81 & 29.95 & 43.99 & 32.84 & 43.19 & 34.91 \\
\bottomrule[1.2pt]
\end{tabular}
\end{table}

\noindent\textbf{Zero-Shot Full-Size Benchmark Results.} We evaluated zero-shot performance on ARC-c, ARC-e, Hellaswag, OpenBookQA \citep{Mihaylov2018obqa}, PIQA \citep{Bisk2020piqa}, SIQA \citep{Sap2019siqa}, WinoGrande and TruthfulQA datasets. The results are shown in Table \ref{table4}. Highest Avg-1 score is achieved by PLDRv5-DAG-5 whereas the highest Avg-2 score is achieved by GPT-Neo-125M. Similar to tinyBenchmarks evaluation, the reference models achieve the best results on ARC-c, ARC-e and Hellaswag and the PLDR-LLMs obtain the highest scores for OpenBookQA, PIQA, SIQA, Winogrande and TruthfulQA.

These trends in benchmark scores in zero and few-shot settings show that DAG-ness of deductive outputs is capable of modifying how well a PLDR-LLM can learn and use the knowlegde acquired from the pretraining dataset for commonsense reasoning and language understanding tasks.

\textbf{Qualitative Results.} We generated 256 tokens as continuation of input text to PLDRv5-1 and PLDRv5-DAG-3 LLMs. The input text was several sentences from the beginning of samples in IMDB Review dataset \citep{Maas2011imdb}. We used nucleus sampling \citep{Holtzman2020nucleus} with $\text{top-p}=0.8$. The model either generates 256 tokens or stops when it encounters an end of sentence ("[END]") token. The input and generated continuation texts are shown in Tables \ref{table5}-\ref{table10} in the appendix. The generated texts suggest that PLDR-LLMs are also susceptible to hallucinations \citep{Ji2023hallucination}.

\noindent\textbf{Ablation Studies.} The loss/accuracy and DAG loss of $\tA_{LM}$ are shown in Fig. \ref{fig3} for a base model and models trained at a lower learning rate, with different tokenizer and longer warm-up steps. The models with "tab" labels are pretrained with a tokenizer model that was trained on the same dataset with same parameters and differs due to random sampling during training. The loss/accuracy curves are in good agreement for PLDR-LLMs pretrained using different tokenizer models and longer warm-up steps. The model PLDRv5-ab-1 with low maximum learning rate converges to a lower loss value and higher accuracy although it does not generate semantically or grammatically meaningful output. This suggests that the model is overfit and learns spurious patterns at lower learning rates and starts to generalize when annealed at high enough learning rates.

The DAG losses of $\tA_{LM}$ for LLMs with different tokenizer model show large differences compared to base model, with PLDRv5-tab-1 exhibiting overflow for most of the pretraining until the very end (Fig.  \ref{fig3e}). Increasing the warm-up steps for PLDRv5-tab-2 brings the DAG loss out of overflow for more pretraining steps. The DAG loss of low learning rate model is higher than the base model with high learning rate and trends in a similar manner.

The benchmark evaluations of the PLDR-LLMs with different tokenizer model and longer warm-up step show a narrow variation in scores and are largely in agreement with base model PLDRv5-2 (Tables \ref{table3} and \ref{table4}). The model with longer warm-up step shows some regression on the average scores.

For both regularized and unregularized PLDR-LLMs, the maximum learning rate and warm-up parameters play an important role during the rest of pretraining. Possible mechanisms as to why this might be the case are that these parameters define how ultra-slow diffusion (random walk in a random potential) takes place over the loss landscape \citep{Hoffer2017diffusion}, the high initial learning rate improves generalization by modifying the learning order of different pattern types \citep{Li2019largelr}, and optimum combination of learning rate and warm-up can avoid or delay formation of high curvature  walls on the loss landscape \citep{Pascanu2013recurrent}.

\clearpage
\begin{table}[!htb]
\caption{Benchmark evaluation results for full-size datasets with zero-shot setting. HS: Hellaswag, OBQA: OpenBookQA, WG: WinoGrande, TQA: TruthfulQA.}
\label{table4}
\centering
\resizebox{\textwidth}{!}{
\begin{tabular}{c c c c c c c c c c c }
\toprule[1.2pt] 
Model & ARC-c & ARC-e & HS & OBQA & PIQA & SIQA & WG & Avg-1 & TQA & Avg-2 \\ 
\midrule[1.2pt]
Cerebras-GPT-111M & 20.56 & 34.89 & 27.10 & 27.60 & 58.11 & 40.23 & 49.33 & 36.83 & 46.32 & 38.02 \\
\midrule 
GPT2-124M & 22.70 & \textbf{39.48} & 31.14 & 27.20 & 62.51 & 41.15 & 50.59 & 39.25 & 40.69 & 39.43 \\  
\midrule 
GPT-Neo-125M & 23.12 & 39.39 & 30.40 & 26.20 & 62.46 & 42.07 & 50.91 & 39.22 & 45.58 & \textbf{40.02} \\
\midrule 
Pythia-160M & \textbf{24.15} & 38.85 & \textbf{31.46} & 27.40 & 62.24 & 40.38 & 50.28 & 39.25 & 44.06 & 39.85 \\ 
\midrule[1.2pt]  
PLDRv5-1 & 21.93 & 36.83 & 29.17 & 29.40 & 61.81 & 42.02 & 50.51 & 38.81 & 43.81 & 39.43 \\ 
\midrule 
PLDRv5-2 & 22.87 & 36.53 & 29.33 & 27.40 & 62.84 & 42.12 & 51.14 & 38.89 & 43.24 & 39.43 \\
\midrule 
PLDRv5-3 & 23.04 & 37.67 & 30.12 & 28.20 & 62.89 & 41.86 & 50.36 & 39.16 & 42.97 & 39.64 \\ 
\midrule  
PLDRv5-4 & 21.59 & 35.86 & 29.48 & 26.80 & 61.26 & 41.45 & 48.30 & 37.82 & \textbf{49.05} & 39.22 \\ 
\midrule[1.2pt] 
PLDRv9-1 & 22.35 & 37.75 & 29.21 & 26.00 & 62.62 & 42.43 & 48.38 & 38.39 & 44.09 & 39.10 \\ 
\midrule 
PLDRv9-2 & 22.44 & 38.68 & 29.63 & 27.40 & 62.84 & 43.09 & 50.28 & 39.19 & 43.40 & 39.72 \\ 
\midrule[1.2pt]
PLDRv5-DAG-1 & 22.27 & 36.62 & 28.99 & 27.20 & 61.75 & 42.58 & 50.04 & 38.49 & 44.12 & 39.20\\ 
\midrule
PLDRv5-DAG-2 & 22.78 & 37.29 & 29.44 & 27.20 & 62.51 & 42.63 & \textbf{51.62} & 39.07 & 43.41 & 39.61 \\
\midrule
PLDRv5-DAG-3 & 21.76 & 38.13 & 29.48 & 26.80 & 61.59 & \textbf{43.50} & 50.36 & 38.80 & 42.97 & 39.32 \\
\midrule
PLDRv5-DAG-4 & 22.27 & 37.75 & 29.46 & 26.60 & 61.92 & 42.48 & 50.51 & 38.71 & 43.50 & 39.31 \\
\midrule
PLDRv5-DAG-5 & 22.01 & 36.78 & 29.62 & \textbf{29.80} & \textbf{63.44} & 42.02 & 51.14 & \textbf{39.26} & 44.24 & 39.88 \\
\midrule
PLDRv5-DAG-6 & 21.08 & 38.26 & 29.45 & 28.20 & 63.00 & 42.43 & 50.20 & 38.94 & 45.10 & 39.71 \\
\midrule
PLDRv5-DAG-7 & 22.70 & 38.80 & 29.21 & 26.60 & 62.13 & 42.84 & 50.12 & 38.91 & 41.91 & 39.29 \\
\midrule[1.2pt]
PLDR-tab-1 & 21.50 & 37.46 & 29.57 & 28.20 & 62.46 & 42.68 & 50.91 & 38.97 & 42.02 & 39.35 \\
\midrule
PLDR-tab-2 & 22.44 & 36.62 & 29.38 & 27.60 & 62.02 & 42.12 & 50.83 & 38.71 & 43.99 & 39.37 \\
\bottomrule[1.2pt]
\end{tabular} } 
\end{table}

\section{Conclusion}

We presented the Large Language Models from Power Law Decoder Representations (PLDR-LLM), a new LLM architecture that utilizes the power law graph attention and has well-defined deductive and inductive outputs. PLDR-LLMs pretrained with $\sim$8B tokens from the RefinedWeb dataset and a batch size of 32 showed competitive performance in zero-shot and few-shot benchmark settings compared to reference scaled dot-product LLMs of similar model size. We studied the DAG loss applied on deductive outputs both as a metric and as a regularizer to observe model characteristics and improve model performance. While training loss and accuracy may not discriminate well between similarly sized LLMs utilizing different hyperparameters and tokenizer models, DAG loss of deductive outputs can provide additional information on model characteristics during pretraining. With a power law graph attention mechanism that leverages both non-linear and linear transformations, well-defined deductive outputs for model characterization and regularization, and competitive performance on benchmarks; PLDR-LLMs define a new class of large language models, diversifying the availability of LLMs to be used in a wide range of NLP applications.

\section*{Acknowledgments}

I am grateful to my parents for their support and patience. This research was conducted independently without support from a grant or corporation.       

\clearpage
\bibliographystyle{unsrtnat}
\bibliography{pldr-llm-references}

\begin{thebibliography}{44}
\providecommand{\natexlab}[1]{#1}
\providecommand{\url}[1]{\texttt{#1}}
\expandafter\ifx\csname urlstyle\endcsname\relax
  \providecommand{\doi}[1]{doi: #1}\else
  \providecommand{\doi}{doi: \begingroup \urlstyle{rm}\Url}\fi

\bibitem[Radford et~al.(2018)Radford, Narasimhan, Salimans, and
  Sutskever]{Radford2018gpt1}
Alec Radford, Karthik Narasimhan, Tim Salimans, and Ilya Sutskever.
\newblock Improving language understanding by generative pre-training.
\newblock 2018.
\newblock URL
  \url{https://cdn.openai.com/research-covers/language-unsupervised/language_understanding_paper.pdf}.

\bibitem[Radford et~al.(2019)Radford, Wu, Child, Luan, Amodei, and
  Sutskever]{Radford2019gpt2}
Alec Radford, Jeff Wu, Rewon Child, David Luan, Dario Amodei, and Ilya
  Sutskever.
\newblock Language models are unsupervised multitask learners.
\newblock 2019.
\newblock URL
  \url{https://cdn.openai.com/better-language-models/language_models_are_unsupervised_multitask_learners.pdf}.

\bibitem[Brown et~al.(2020)Brown, Mann, Ryder, Subbiah, Kaplan, Dhariwal,
  Neelakantan, Shyam, Sastry, Askell, Agarwal, Herbert-Voss, Krueger, Henighan,
  Child, Ramesh, Ziegler, Wu, Winter, Hesse, Chen, Sigler, Litwin, Gray, Chess,
  Clark, Berner, McCandlish, Radford, Sutskever, and Amodei]{Brown2020gpt3}
Tom~B. Brown, Benjamin Mann, Nick Ryder, Melanie Subbiah, Jared Kaplan,
  Prafulla Dhariwal, Arvind Neelakantan, Pranav Shyam, Girish Sastry, Amanda
  Askell, Sandhini Agarwal, Ariel Herbert-Voss, Gretchen Krueger, Tom Henighan,
  Rewon Child, Aditya Ramesh, Daniel~M. Ziegler, Jeffrey Wu, Clemens Winter,
  Christopher Hesse, Mark Chen, Eric Sigler, Mateusz Litwin, Scott Gray,
  Benjamin Chess, Jack Clark, Christopher Berner, Sam McCandlish, Alec Radford,
  Ilya Sutskever, and Dario Amodei.
\newblock Language models are few-shot learners.
\newblock NIPS '20, Red Hook, NY, USA, 2020. Curran Associates Inc.
\newblock ISBN 9781713829546.

\bibitem[Touvron et~al.(2023{\natexlab{a}})Touvron, Lavril, Izacard, Martinet,
  Lachaux, Lacroix, Rozi{\`e}re, Goyal, Hambro, Azhar, Rodriguez, Joulin,
  Grave, and Lample]{Touvron2023llama}
Hugo Touvron, Thibaut Lavril, Gautier Izacard, Xavier Martinet, Marie-Anne
  Lachaux, Timoth{\'e}e Lacroix, Baptiste Rozi{\`e}re, Naman Goyal, Eric
  Hambro, Faisal Azhar, Aurelien Rodriguez, Armand Joulin, Edouard Grave, and
  Guillaume Lample.
\newblock Llama: Open and efficient foundation language models.
\newblock \emph{ArXiv}, abs/2302.13971, 2023{\natexlab{a}}.

\bibitem[Touvron et~al.(2023{\natexlab{b}})Touvron, Martin, Stone, Albert,
  Almahairi, Babaei, Bashlykov, Batra, Bhargava, Bhosale, Bikel, Blecher,
  Ferrer, Chen, Cucurull, Esiobu, Fernandes, Fu, Fu, Fuller, Gao, Goswami,
  Goyal, Hartshorn, Hosseini, Hou, Inan, Kardas, Kerkez, Khabsa, Kloumann,
  Korenev, Koura, Lachaux, Lavril, Lee, Liskovich, Lu, Mao, Martinet, Mihaylov,
  Mishra, Molybog, Nie, Poulton, Reizenstein, Rungta, Saladi, Schelten, Silva,
  Smith, Subramanian, Tan, Tang, Taylor, Williams, Kuan, Xu, Yan, Zarov, Zhang,
  Fan, Kambadur, Narang, Rodriguez, Stojnic, Edunov, and
  Scialom]{Touvron2023llama2}
Hugo Touvron, Louis Martin, Kevin~R. Stone, Peter Albert, Amjad Almahairi,
  Yasmine Babaei, Nikolay Bashlykov, Soumya Batra, Prajjwal Bhargava, Shruti
  Bhosale, Daniel~M. Bikel, Lukas Blecher, Cristian~Cant{\'o}n Ferrer, Moya
  Chen, Guillem Cucurull, David Esiobu, Jude Fernandes, Jeremy Fu, Wenyin Fu,
  Brian Fuller, Cynthia Gao, Vedanuj Goswami, Naman Goyal, Anthony~S.
  Hartshorn, Saghar Hosseini, Rui Hou, Hakan Inan, Marcin Kardas, Viktor
  Kerkez, Madian Khabsa, Isabel~M. Kloumann, A.~V. Korenev, Punit~Singh Koura,
  Marie-Anne Lachaux, Thibaut Lavril, Jenya Lee, Diana Liskovich, Yinghai Lu,
  Yuning Mao, Xavier Martinet, Todor Mihaylov, Pushkar Mishra, Igor Molybog,
  Yixin Nie, Andrew Poulton, Jeremy Reizenstein, Rashi Rungta, Kalyan Saladi,
  Alan Schelten, Ruan Silva, Eric~Michael Smith, R.~Subramanian, Xia Tan, Binh
  Tang, Ross Taylor, Adina Williams, Jian~Xiang Kuan, Puxin Xu, Zhengxu Yan,
  Iliyan Zarov, Yuchen Zhang, Angela Fan, Melanie Kambadur, Sharan Narang,
  Aurelien Rodriguez, Robert Stojnic, Sergey Edunov, and Thomas Scialom.
\newblock Llama 2: Open foundation and fine-tuned chat models.
\newblock \emph{ArXiv}, abs/2307.09288, 2023{\natexlab{b}}.

\bibitem[Kaplan et~al.(2020)Kaplan, McCandlish, Henighan, Brown, Chess, Child,
  Gray, Radford, Wu, and Amodei]{Kaplan2020scaling}
Jared Kaplan, Sam McCandlish, Tom Henighan, Tom~B Brown, Benjamin Chess, Rewon
  Child, Scott Gray, Alec Radford, Jeffrey Wu, and Dario Amodei.
\newblock Scaling laws for neural language models.
\newblock \emph{arXiv preprint arXiv:2001.08361}, 2020.

\bibitem[Hoffmann et~al.(2022)Hoffmann, Borgeaud, Mensch, Buchatskaya, Cai,
  Rutherford, de~Las~Casas, Hendricks, Welbl, Clark, Hennigan, Noland,
  Millican, van~den Driessche, Damoc, Guy, Osindero, Simonyan, Elsen, Vinyals,
  Rae, and Sifre]{Hoffmann2022chinchilla}
Jordan Hoffmann, Sebastian Borgeaud, Arthur Mensch, Elena Buchatskaya, Trevor
  Cai, Eliza Rutherford, Diego de~Las~Casas, Lisa~Anne Hendricks, Johannes
  Welbl, Aidan Clark, Tom Hennigan, Eric Noland, Katie Millican, George van~den
  Driessche, Bogdan Damoc, Aurelia Guy, Simon Osindero, Karen Simonyan, Erich
  Elsen, Oriol Vinyals, Jack~W. Rae, and Laurent Sifre.
\newblock Training compute-optimal large language models.
\newblock In \emph{Proceedings of the 36th International Conference on Neural
  Information Processing Systems}, NIPS '22, Red Hook, NY, USA, 2022. Curran
  Associates Inc.
\newblock ISBN 9781713871088.

\bibitem[Penedo et~al.(2023)Penedo, Malartic, Hesslow, Cojocaru, Alobeidli,
  Cappelli, Pannier, Almazrouei, and Launay]{Penedo2023falcon}
Guilherme Penedo, Quentin Malartic, Daniel Hesslow, Ruxandra Cojocaru, Hamza
  Alobeidli, Alessandro Cappelli, Baptiste Pannier, Ebtesam Almazrouei, and
  Julien Launay.
\newblock The refinedweb dataset for falcon llm: outperforming curated corpora
  with web data only.
\newblock In \emph{Proceedings of the 37th International Conference on Neural
  Information Processing Systems}, NIPS '23, Red Hook, NY, USA, 2023. Curran
  Associates Inc.

\bibitem[Gao et~al.(2020)Gao, Biderman, Black, Golding, Hoppe, Foster, Phang,
  He, Thite, Nabeshima, et~al.]{Gao2020pile}
Leo Gao, Stella Biderman, Sid Black, Laurence Golding, Travis Hoppe, Charles
  Foster, Jason Phang, Horace He, Anish Thite, Noa Nabeshima, et~al.
\newblock The pile: An 800gb dataset of diverse text for language modeling.
\newblock \emph{arXiv preprint arXiv:2101.00027}, 2020.

\bibitem[Liu et~al.(2024)Liu, Zhao, Iandola, Lai, Tian, Fedorov, Xiong, Chang,
  Shi, Krishnamoorthi, et~al.]{Liu2024mobilellm}
Zechun Liu, Changsheng Zhao, Forrest Iandola, Chen Lai, Yuandong Tian, Igor
  Fedorov, Yunyang Xiong, Ernie Chang, Yangyang Shi, Raghuraman Krishnamoorthi,
  et~al.
\newblock Mobilellm: Optimizing sub-billion parameter language models for
  on-device use cases.
\newblock \emph{arXiv preprint arXiv:2402.14905}, 2024.

\bibitem[Vaswani et~al.(2017)Vaswani, Shazeer, Parmar, Uszkoreit, Jones, Gomez,
  Kaiser, and Polosukhin]{Vasvani2017}
Ashish Vaswani, Noam Shazeer, Niki Parmar, Jakob Uszkoreit, Llion Jones,
  Aidan~N. Gomez, undefinedukasz Kaiser, and Illia Polosukhin.
\newblock Attention is all you need.
\newblock In \emph{Proceedings of the 31st International Conference on Neural
  Information Processing Systems}, NIPS'17, pages 6000--6010, Red Hook, NY,
  USA, 2017. Curran Associates Inc.
\newblock ISBN 9781510860964.

\bibitem[Gokden(2021)]{Gokden2021}
Burc Gokden.
\newblock Power law graph transformer for machine translation and
  representation learning.
\newblock \emph{arXiv preprint arXiv:2107.02039}, 2021.

\bibitem[Gokden(2019)]{Gokden2019}
Burc Gokden.
\newblock Coulgat: An experiment on interpretability of graph attention
  networks.
\newblock \emph{arXiv preprint arXiv:1912.08409}, 2019.

\bibitem[Zheng et~al.(2018)Zheng, Aragam, Ravikumar, and
  Xing]{Zheng2018notears}
Xun Zheng, Bryon Aragam, Pradeep Ravikumar, and Eric~P. Xing.
\newblock Dags with no tears: continuous optimization for structure learning.
\newblock In \emph{Proceedings of the 32nd International Conference on Neural
  Information Processing Systems}, NIPS'18, page 9492–9503, Red Hook, NY,
  USA, 2018. Curran Associates Inc.

\bibitem[Shazeer(2020)]{Shazeer2020swiglu}
Noam Shazeer.
\newblock Glu variants improve transformer.
\newblock \emph{arXiv preprint arXiv:2002.05202}, 2020.

\bibitem[Dauphin et~al.(2017)Dauphin, Fan, Auli, and Grangier]{Dauphin2017glu}
Yann~N. Dauphin, Angela Fan, Michael Auli, and David Grangier.
\newblock Language modeling with gated convolutional networks.
\newblock In \emph{Proceedings of the 34th International Conference on Machine
  Learning - Volume 70}, ICML'17, page 933–941. JMLR.org, 2017.

\bibitem[Ramachandran et~al.(2017)Ramachandran, Zoph, and
  Le]{Ramachandran2017swish}
Prajit Ramachandran, Barret Zoph, and Quoc~V Le.
\newblock Searching for activation functions.
\newblock \emph{arXiv preprint arXiv:1710.05941}, 2017.

\bibitem[Elfwing et~al.(2018)Elfwing, Uchibe, and Doya]{Elfwing2018silu}
Stefan Elfwing, Eiji Uchibe, and Kenji Doya.
\newblock Sigmoid-weighted linear units for neural network function
  approximation in reinforcement learning.
\newblock \emph{Neural networks}, 107:\penalty0 3--11, 2018.

\bibitem[Su et~al.(2021)Su, Lu, Pan, Wen, and Liu]{Su2021roformer}
Jianlin Su, Yu~Lu, Shengfeng Pan, Bo~Wen, and Yunfeng Liu.
\newblock Roformer: Enhanced transformer with rotary position embedding, 2021.

\bibitem[Ba(2016)]{Ba2016layer}
Jimmy~Lei Ba.
\newblock Layer normalization.
\newblock \emph{arXiv preprint arXiv:1607.06450}, 2016.

\bibitem[Loshchilov and Hutter(2019)]{Loshchilov2019adamw}
Ilya Loshchilov and Frank Hutter.
\newblock Decoupled weight decay regularization.
\newblock In \emph{7th International Conference on Learning Representations,
  {ICLR} 2019, New Orleans, LA, USA, May 6-9, 2019}. OpenReview.net, 2019.

\bibitem[Loshchilov and Hutter(2016)]{Loshchilov2016sgdr}
Ilya Loshchilov and Frank Hutter.
\newblock Sgdr: Stochastic gradient descent with warm restarts.
\newblock \emph{arXiv preprint arXiv:1608.03983}, 2016.

\bibitem[Kudo and Richardson(2018)]{Kudo2018sentencepiece}
Taku Kudo and John Richardson.
\newblock {S}entence{P}iece: A simple and language independent subword
  tokenizer and detokenizer for neural text processing.
\newblock In \emph{Proceedings of the 2018 Conference on Empirical Methods in
  Natural Language Processing: System Demonstrations}, pages 66--71, Brussels,
  Belgium, November 2018. Association for Computational Linguistics.
\newblock \doi{10.18653/v1/D18-2012}.
\newblock URL \url{https://aclanthology.org/D18-2012}.

\bibitem[Kudo(2018)]{Kudo2018subword}
Taku Kudo.
\newblock Subword regularization: Improving neural network translation models
  with multiple subword candidates.
\newblock In Iryna Gurevych and Yusuke Miyao, editors, \emph{Proceedings of the
  56th Annual Meeting of the Association for Computational Linguistics (Volume
  1: Long Papers)}, pages 66--75, Melbourne, Australia, July 2018. Association
  for Computational Linguistics.
\newblock \doi{10.18653/v1/P18-1007}.
\newblock URL \url{https://aclanthology.org/P18-1007}.

\bibitem[Maia~Polo et~al.(2024)Maia~Polo, Weber, Choshen, Sun, Xu, and
  Yurochkin]{Polo2024tinybenchmarks}
Felipe Maia~Polo, Lucas Weber, Leshem Choshen, Yuekai Sun, Gongjun Xu, and
  Mikhail Yurochkin.
\newblock tinybenchmarks: evaluating llms with fewer examples.
\newblock \emph{arXiv preprint arXiv:2402.14992}, 2024.

\bibitem[Gao et~al.(2024)Gao, Tow, Abbasi, Biderman, Black, DiPofi, Foster,
  Golding, Hsu, Le~Noac'h, Li, McDonell, Muennighoff, Ociepa, Phang, Reynolds,
  Schoelkopf, Skowron, Sutawika, Tang, Thite, Wang, Wang, and Zou]{evalharness}
Leo Gao, Jonathan Tow, Baber Abbasi, Stella Biderman, Sid Black, Anthony
  DiPofi, Charles Foster, Laurence Golding, Jeffrey Hsu, Alain Le~Noac'h,
  Haonan Li, Kyle McDonell, Niklas Muennighoff, Chris Ociepa, Jason Phang,
  Laria Reynolds, Hailey Schoelkopf, Aviya Skowron, Lintang Sutawika, Eric
  Tang, Anish Thite, Ben Wang, Kevin Wang, and Andy Zou.
\newblock A framework for few-shot language model evaluation, 07 2024.
\newblock URL \url{https://zenodo.org/records/12608602}.

\bibitem[Dey et~al.(2023)Dey, Gosal, Khachane, Marshall, Pathria, Tom,
  Hestness, et~al.]{dey2023cerebras}
Nolan Dey, Gurpreet Gosal, Hemant Khachane, William Marshall, Ribhu Pathria,
  Marvin Tom, Joel Hestness, et~al.
\newblock Cerebras-gpt: Open compute-optimal language models trained on the
  cerebras wafer-scale cluster.
\newblock \emph{arXiv preprint arXiv:2304.03208}, 2023.

\bibitem[Black et~al.(2021)Black, Leo, Wang, Leahy, and Biderman]{gptneo}
Sid Black, Gao Leo, Phil Wang, Connor Leahy, and Stella Biderman.
\newblock {GPT-Neo: Large Scale Autoregressive Language Modeling with
  Mesh-Tensorflow}, March 2021.
\newblock URL \url{https://doi.org/10.5281/zenodo.5297715}.

\bibitem[Biderman et~al.(2023)Biderman, Schoelkopf, Anthony, Bradley, O'Brien,
  Hallahan, Khan, Purohit, Prashanth, Raff, Skowron, Sutawika, and Van
  Der~Wal]{Biderman2023pythia}
Stella Biderman, Hailey Schoelkopf, Quentin Anthony, Herbie Bradley, Kyle
  O'Brien, Eric Hallahan, Mohammad~Aflah Khan, Shivanshu Purohit, USVSN~Sai
  Prashanth, Edward Raff, Aviya Skowron, Lintang Sutawika, and Oskar Van
  Der~Wal.
\newblock Pythia: a suite for analyzing large language models across training
  and scaling.
\newblock In \emph{Proceedings of the 40th International Conference on Machine
  Learning}, ICML'23. JMLR.org, 2023.

\bibitem[Lin et~al.(2022)Lin, Hilton, and Evans]{Lin2021truthfulqa}
Stephanie Lin, Jacob Hilton, and Owain Evans.
\newblock {T}ruthful{QA}: Measuring how models mimic human falsehoods.
\newblock In \emph{Proceedings of the 60th Annual Meeting of the Association
  for Computational Linguistics (Volume 1: Long Papers)}, pages 3214--3252,
  Dublin, Ireland, May 2022. Association for Computational Linguistics.
\newblock \doi{10.18653/v1/2022.acl-long.229}.
\newblock URL \url{https://aclanthology.org/2022.acl-long.229}.

\bibitem[Clark et~al.(2018)Clark, Cowhey, Etzioni, Khot, Sabharwal, Schoenick,
  and Tafjord]{Clark2018arc}
Peter Clark, Isaac Cowhey, Oren Etzioni, Tushar Khot, Ashish Sabharwal, Carissa
  Schoenick, and Oyvind Tafjord.
\newblock Think you have solved question answering? try arc, the ai2 reasoning
  challenge.
\newblock \emph{arXiv:1803.05457v1}, 2018.

\bibitem[Zellers et~al.(2019)Zellers, Holtzman, Bisk, Farhadi, and
  Choi]{Zellers2019hellaswag}
Rowan Zellers, Ari Holtzman, Yonatan Bisk, Ali Farhadi, and Yejin Choi.
\newblock Hellaswag: Can a machine really finish your sentence?
\newblock In \emph{Proceedings of the 57th Annual Meeting of the Association
  for Computational Linguistics}, 2019.

\bibitem[Hendrycks et~al.(2021{\natexlab{a}})Hendrycks, Burns, Basart, Zou,
  Mazeika, Song, and Steinhardt]{Hendrycks2021test}
Dan Hendrycks, Collin Burns, Steven Basart, Andy Zou, Mantas Mazeika, Dawn
  Song, and Jacob Steinhardt.
\newblock Measuring massive multitask language understanding.
\newblock \emph{Proceedings of the International Conference on Learning
  Representations (ICLR)}, 2021{\natexlab{a}}.

\bibitem[Hendrycks et~al.(2021{\natexlab{b}})Hendrycks, Burns, Basart, Critch,
  Li, Song, and Steinhardt]{Hendrycks2021ethics}
Dan Hendrycks, Collin Burns, Steven Basart, Andrew Critch, Jerry Li, Dawn Song,
  and Jacob Steinhardt.
\newblock Aligning ai with shared human values.
\newblock \emph{Proceedings of the International Conference on Learning
  Representations (ICLR)}, 2021{\natexlab{b}}.

\bibitem[Sakaguchi et~al.(2021)Sakaguchi, Bras, Bhagavatula, and
  Choi]{Keisuke2019winogrande}
Keisuke Sakaguchi, Ronan~Le Bras, Chandra Bhagavatula, and Yejin Choi.
\newblock Winogrande: an adversarial winograd schema challenge at scale.
\newblock \emph{Commun. ACM}, 64\penalty0 (9):\penalty0 99–106, August 2021.
\newblock ISSN 0001-0782.
\newblock \doi{10.1145/3474381}.
\newblock URL \url{https://doi.org/10.1145/3474381}.

\bibitem[Mihaylov et~al.(2018)Mihaylov, Clark, Khot, and
  Sabharwal]{Mihaylov2018obqa}
Todor Mihaylov, Peter Clark, Tushar Khot, and Ashish Sabharwal.
\newblock Can a suit of armor conduct electricity? a new dataset for open book
  question answering.
\newblock In \emph{EMNLP}, 2018.

\bibitem[Bisk et~al.(2020)Bisk, Zellers, Bras, Gao, and Choi]{Bisk2020piqa}
Yonatan Bisk, Rowan Zellers, Ronan~Le Bras, Jianfeng Gao, and Yejin Choi.
\newblock Piqa: Reasoning about physical commonsense in natural language.
\newblock In \emph{Thirty-Fourth AAAI Conference on Artificial Intelligence},
  2020.

\bibitem[Sap et~al.(2019)Sap, Rashkin, Chen, Le~Bras, and Choi]{Sap2019siqa}
Maarten Sap, Hannah Rashkin, Derek Chen, Ronan Le~Bras, and Yejin Choi.
\newblock Social {IQ}a: Commonsense reasoning about social interactions.
\newblock In \emph{Proceedings of the 2019 Conference on Empirical Methods in
  Natural Language Processing and the 9th International Joint Conference on
  Natural Language Processing (EMNLP-IJCNLP)}, pages 4463--4473, Hong Kong,
  China, November 2019. Association for Computational Linguistics.
\newblock \doi{10.18653/v1/D19-1454}.
\newblock URL \url{https://aclanthology.org/D19-1454}.

\bibitem[Maas et~al.(2011)Maas, Daly, Pham, Huang, Ng, and Potts]{Maas2011imdb}
Andrew~L. Maas, Raymond~E. Daly, Peter~T. Pham, Dan Huang, Andrew~Y. Ng, and
  Christopher Potts.
\newblock Learning word vectors for sentiment analysis.
\newblock In \emph{Proceedings of the 49th Annual Meeting of the Association
  for Computational Linguistics: Human Language Technologies}, pages 142--150,
  Portland, Oregon, USA, June 2011. Association for Computational Linguistics.
\newblock URL \url{http://www.aclweb.org/anthology/P11-1015}.

\bibitem[Holtzman et~al.(2020)Holtzman, Buys, Du, Forbes, and
  Choi]{Holtzman2020nucleus}
Ari Holtzman, Jan Buys, Li~Du, Maxwell Forbes, and Yejin Choi.
\newblock The curious case of neural text degeneration.
\newblock In \emph{8th International Conference on Learning Representations,
  {ICLR} 2020, Addis Ababa, Ethiopia, April 26-30, 2020}. OpenReview.net, 2020.
\newblock URL \url{https://openreview.net/forum?id=rygGQyrFvH}.

\bibitem[Ji et~al.(2023)Ji, Lee, Frieske, Yu, Su, Xu, Ishii, Bang, Madotto, and
  Fung]{Ji2023hallucination}
Ziwei Ji, Nayeon Lee, Rita Frieske, Tiezheng Yu, Dan Su, Yan Xu, Etsuko Ishii,
  Ye~Jin Bang, Andrea Madotto, and Pascale Fung.
\newblock Survey of hallucination in natural language generation.
\newblock \emph{ACM Computing Surveys}, 55\penalty0 (12):\penalty0 1–38,
  March 2023.
\newblock ISSN 1557-7341.
\newblock \doi{10.1145/3571730}.
\newblock URL \url{http://dx.doi.org/10.1145/3571730}.

\bibitem[Hoffer et~al.(2017)Hoffer, Hubara, and Soudry]{Hoffer2017diffusion}
Elad Hoffer, Itay Hubara, and Daniel Soudry.
\newblock Train longer, generalize better: closing the generalization gap in
  large batch training of neural networks.
\newblock In \emph{Proceedings of the 31st International Conference on Neural
  Information Processing Systems}, NIPS'17, page 1729–1739, Red Hook, NY,
  USA, 2017. Curran Associates Inc.
\newblock ISBN 9781510860964.

\bibitem[Li et~al.(2019)Li, Wei, and Ma]{Li2019largelr}
Yuanzhi Li, Colin Wei, and Tengyu Ma.
\newblock \emph{Towards explaining the regularization effect of initial large
  learning rate in training neural networks}.
\newblock Curran Associates Inc., Red Hook, NY, USA, 2019.

\bibitem[Pascanu et~al.(2013)Pascanu, Mikolov, and
  Bengio]{Pascanu2013recurrent}
Razvan Pascanu, Tomas Mikolov, and Yoshua Bengio.
\newblock On the difficulty of training recurrent neural networks.
\newblock ICML'13, page III–1310–III–1318. JMLR.org, 2013.

\end{thebibliography}

\clearpage
\appendix

\section*{Appendix}
\section{Model Architecture Diagrams}
\begin{figure}[hb]
\centering
\begin{subfigure}[b]{0.45\textwidth}
	\centering
	\includegraphics[width=0.95\textwidth]{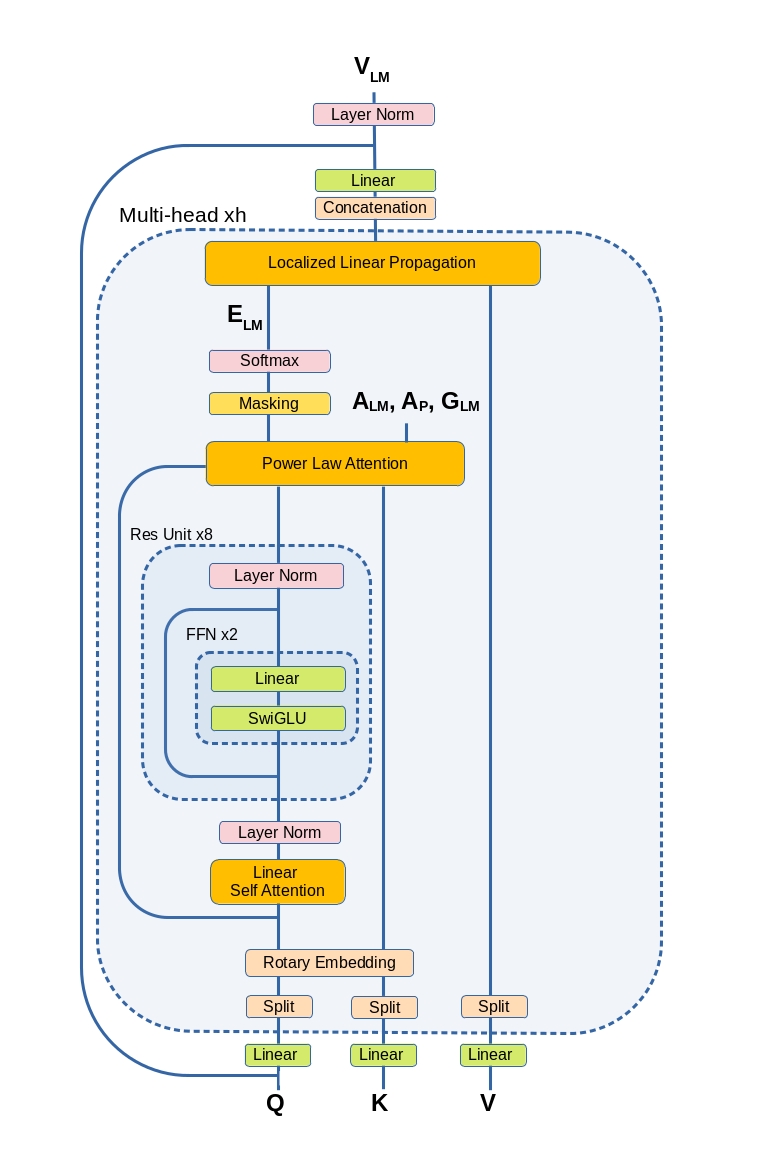}
	\caption{PLDRv5 Multihead Attention}
	\label{fig4a}
\end{subfigure}
\hfill
\begin{subfigure}[b]{0.45\textwidth}
	\centering
	\includegraphics[width=0.95\textwidth]{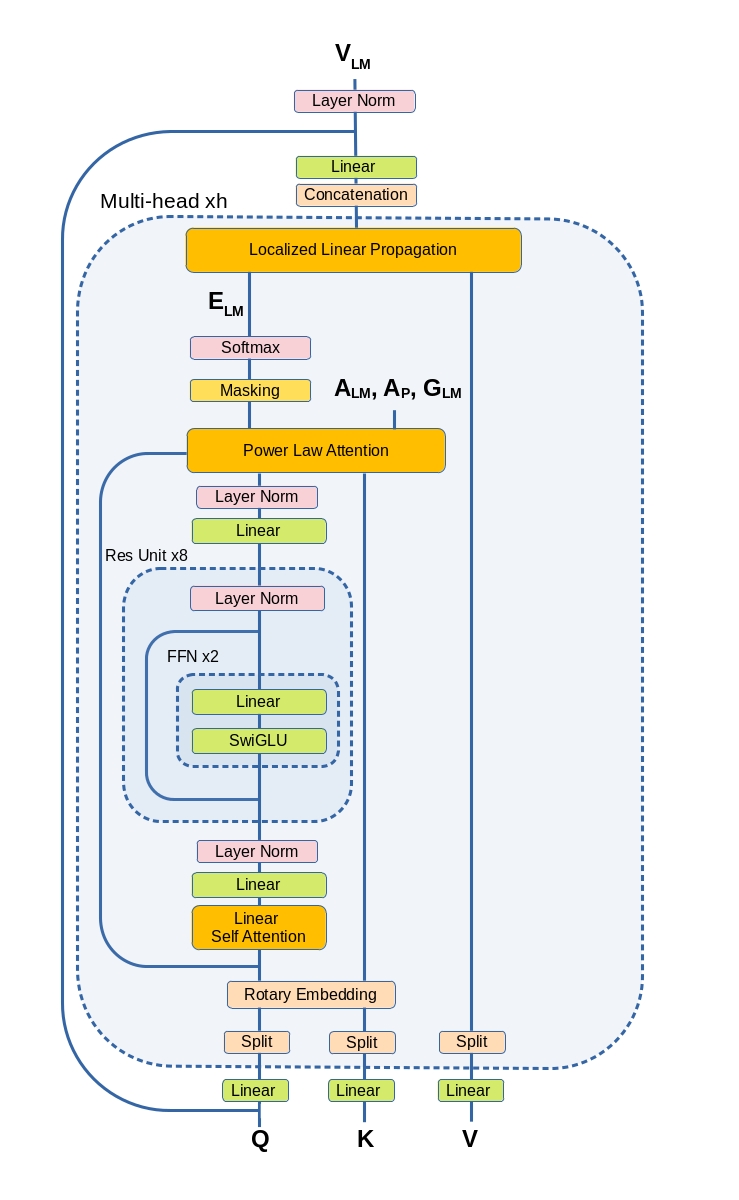}
	\caption{PLDRv9 Multihead Attention}
	\label{fig4b}
\end{subfigure}
\hfill
\centering
\begin{subfigure}[hb]{0.42\textwidth}
	\centering
	\includegraphics[width=0.50\textwidth]{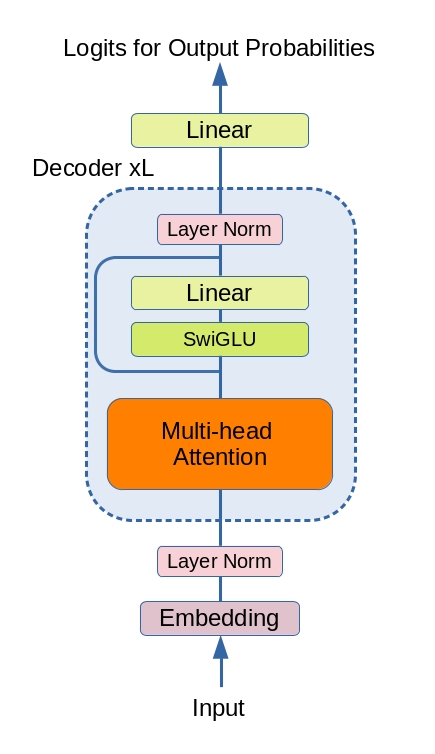}
	\caption{PLDR-LLM Model Architecture}
	\label{fig4c}
\end{subfigure}
\caption{PLDR-LLM model and multihead attention diagrams for PLDRv5 and PLDRv9 designs. PLDRv9 only differs in resizing of layers before and after residual networks for the metric learner. Feedforward network (FFN) is composed of SwiGLU and Linear layers. }
\label{fig4}
\end{figure}

\newpage
\section{Benchmark Datasets}

\textbf{tinyBenchmarks}. tinyBenchmarks consists of curated samples from datasets that are part of popular benchmarks (OpenLLM Leaderboard, MMLU, HELM and AlpacaEval 2.0). It takes advantage of models of educational assessments from psychometrics to reduce the samples to a small fraction of the actual dataset to evaluate performance of LLMs \citep{Polo2024tinybenchmarks}.

\textbf{ARC}. The AI2 Reasoning Challenge (ARC) dataset consists of multiple-choice grade school questions from $3^{rd}$ to $9^{th}$ grade. It consists of an easy set and a challenge set. The challenge set contains the questions answered incorrectly by both a retrieval based algorithm and a word co-occurrence algorithm \citep{Clark2018arc}.

\textbf{Hellaswag}. Harder Endings, Longer contexts, and Low-shot Activities for Situations With Adversarial Generations dataset is a commonsense natural language inference dataset that was prepared using adversial filtering to create problems that are challenging to models, yet easy for humans \citep{Zellers2019hellaswag}.

\textbf{MMLU}. Massive Multitask Language Understanding is a multiple-choice benchmark that covers 57 tasks (e.g. elementary mathematics, US history, computer science, etc.). It aims to how well models can apply knowledge learned during pretraining \citep{Hendrycks2021test, Hendrycks2021ethics}.

\textbf{WinoGrande}. WinoGrande is a more challenging version of Winograd Schema Challenge that is a commonsense reasoning benchmark based on a set of pronoun resolution problems designed to be unsolvable for statistical models that rely on selectional preferences or word associations \citep{Keisuke2019winogrande}.

\textbf{TruthfulQA}. TruthfulQA is a benchmark that aims to measure truthfullness of a model. It consists of questions covering 38 categories such as health, law, finance and politics. The model should avoid imitating human contexts in pretraining dataset to perform well, since the questions are selected from the ones humans would answer incorrectly due to a false belief or misconception \citep{Lin2021truthfulqa}.

\textbf{OpenBookQA}. OpenBookQA is a question answering dataset that consists of about 6000 questions accompanied with scientific facts. To answer the questions correctly the model needs to combine with extra common knowledge beyond the facts included in the dataset \citep{Mihaylov2018obqa}.

\textbf{PIQA}. Physical Interaction:Question Answering dataset is a physical commonsense benchmark that aims to evaluate model performance for concepts that are traditionally only seen or experienced in the real world \citep{Bisk2020piqa}.

\textbf{SIQA}. Social Intelligence QA dataset is a social commonsense reasoning benchmark that aims to evaluate model performance for social situations. It consists of 38000 multiple-choice questions for probing emotional and social intelligence in a variety of everyday situations \citep{Sap2019siqa}.

\textbf{IMDB Review}. IMDB Review dataset is a collection of 50000 reviews with each movie having no more than 30 reviews. It was compiled for sentiment analysis and consists of an even number of highly polarized negative ($\leq4$ out of $10$) and positive ($\geq7$ out of $10$) reviews \citep{Maas2011imdb}.

\newpage
\section{Sample Text Outputs from PLDR-LLM}

\begin{table}[!htb]
\caption{Input is several sentences from the beginning of a review sample from IMDB Review dataset. Continuation is the generated text output from PLDRv5-1 LLM model with nucleus sampling at $\text{top-p}=0.8$. The model generates text for 256 tokens or until it encounters an end of sentence ("[END]") token. }
\label{table5}
\centering
\resizebox{\textwidth}{!}{

\begin{tabular}{| p{5cm} | p{5cm} | }
\hline 
\textbf{Input} & \textbf{Continuation} \\ 
\hline
I love sci-fi and am willing to put up with a lot. Sci-fi movies/TV are usually underfunded, under-appreciated and misunderstood. I tried to like this, I really did, but it is to good TV sci-fi as Babylon 5 is to Star Trek (the original). Silly prosthetics, cheap cardboard sets, stilted dialogues, CG that doesn't match the background, and painfully one-dimensional characters cannot be overcome with a 'sci-fi' setting. 
&
And here we are, in the midst of the constraints of a man and a woman, battling against those beautiful arms of a fantastic character, who staggers in their various adventures of pain and suffering, to the plotline of an apocalyptic ideal of a film or TV show or fantasy/monster fantasy, from the rest of the set to the list of "scary bones" that needs to be dealt with. Are there any so-called sci-fi franchises that might not survive a modern such franchise, either? I don't know. There are hundreds of millions of us who don't read the YA novels. There are thousands of them in the stands, not long after we get to enjoy them, either at their own terms or at some point at their own expense. 1999, for example, I've heard the name, and I've seen one or two similar YA series, because they don't have a good movie, but I've seen two. I haven't seen a single horror or thriller in the series. In that time, the genre is going to be a phenomenon of authors that will be put on the page and will never be complete. You know, you know, they
\\ \hline
\end{tabular} } 
\end{table}

\begin{table}[!htb]
\caption{Input is several sentences from the beginning of a review sample from IMDB Review dataset. Continuation is the generated text output from PLDRv5-1 LLM model with nucleus sampling at $\text{top-p}=0.8$. The model generates text for 256 tokens or until it encounters an end of sentence ("[END]") token.}
\label{table6}
\centering
\resizebox{\textwidth}{!}{

\begin{tabular}{| p{5cm} | p{5cm} | }
\hline 
\textbf{Input} & \textbf{Continuation} \\ 
\hline
"Cover Girl" is a lacklustre WWII musical with absolutely nothing memorable about it, save for its signature song, "Long Ago and Far Away." This film came out before Gene Kelly really hit his artistic stride, and while there are evidences of his burgeoning talent here, mostly he plays sidekick to Rita Hayworth. And there's the problem. Rita Hayworth is gorgeous, no doubt about that. But she's simply not a compelling screen presence.
&
Her voice is hot and sultry, her story is bumbling and raw. She's become an immense character in the way she was almost four decades ago. As much as her films have failed at making any sort of punchy fare, there's not one who can put her down for the most part. As someone who recently turned down the idea of making a documentary about her adventures, I'm shocked. In 2018, my friend Dracut Qenisa expressed to me that I wanted to do a film about an only child who has become very passionate about sports. This film is about those things. We aren't just discussing the underworld. But it's about the amazing things we do every day, that shape our lives, and in so many ways inspire us to do the same. And then we find the precious in our own lives. And we find the pure beauty of the rest of the world. (See: The Woman in the Metal, Is Out with 'Insomnia') And when we are about to make our own film, we can find the perfect actor to meet the kind of aesthetic our own skin can create. And that's why I wanted to make this film. My experiences in
\\ \hline
\end{tabular} } 
\end{table}

\begin{table}[!htb]
\caption{Input is several sentences from the beginning of a review sample from IMDB Review dataset. Continuation is the generated text output from PLDRv5-1 LLM model with nucleus sampling at $\text{top-p}=0.8$. The model generates text for 256 tokens or until it encounters an end of sentence ("[END]") token.}
\label{table7}
\centering
\resizebox{\textwidth}{!}{

\begin{tabular}{| p{5cm} | p{5cm} | }
\hline 
\textbf{Input} & \textbf{Continuation} \\ 
\hline
No emotion. Bad music (and I am a reformed eighties metal guy, so I would be vulnerable to some good stuff.) Everything is half done. The brother is a parody, there is nothing to hold us in except to see what horror from the eighties they'll pull up next. The tour manager tries to bring a humanity to the thing but isn't given enough time.
&
As in the most depressing thing on his tour, he is using something, the adrenaline of it, that many fans have never had before. As I mentioned, the audience knows what this album is about. They know the history of the band, they know that there's a new album coming out on the road this summer. There's a rap, it's an underrated musical number, and the band's sounds to us are rooted in it. It's an interesting story about the original album that never was, with its beginnings as a two-man band, and a page-turner. It's what we heard the last 30 years before, which is another horror band. But now the band are ready to move on, and they're closing the tour with an out-of-date set of music. That means you don't even have to be familiar with the band or what they're doing, so there's no arguing here. The band doesn't know what the album is about, but they have a nice audience to play, and they're the kind of people who have been to the band when they're only going to band one, and it's about the weight of
\\ \hline
\end{tabular} } 
\end{table}

\begin{table}[!htb]
\caption{Input is several sentences from the beginning of a review sample from IMDB Review dataset. Continuation is the generated text output from PLDRv5-DAG-3 LLM model with nucleus sampling at $\text{top-p}=0.8$. The model generates text for 256 tokens or until it encounters an end of sentence ("[END]") token. }
\label{table8}
\centering
\resizebox{\textwidth}{!}{

\begin{tabular}{| p{5cm} | p{5cm} | }
\hline 
\textbf{Input} & \textbf{Continuation} \\ 
\hline
I love sci-fi and am willing to put up with a lot. Sci-fi movies/TV are usually underfunded, under-appreciated and misunderstood. I tried to like this, I really did, but it is to good TV sci-fi as Babylon 5 is to Star Trek (the original). Silly prosthetics, cheap cardboard sets, stilted dialogues, CG that doesn't match the background, and painfully one-dimensional characters cannot be overcome with a 'sci-fi' setting. 
&
And I mean the costumes are a total reflection of the film. Science Fiction is good, well done, right? The feature film characters don't fit into the big picture, I love what they do to that genre. Not so much the supernatural/sustainable/soldier/etc. scene, but it's just the sort of perspective that will make you take off and see what you like. Well done! It's a shame that we have to give it our all, too. But the dialogue is worth it. Some of the props are ridiculous, but it's all because they're going to make you think about what the story is, which one to use. Like, 20 years ago, when I first seen this film, I thought it was better. Now, I know this is one of the things that was discussed in the film. I'm not sure if the script really ran, I think it's on point. The actors are so much the same, they all still have their roles, but they all still have their names. The actors have their roots in the fiction, the script has their roots in the history of the film, and the stories have their roots in the storytelling. But,
\\ \hline
\end{tabular} } 
\end{table}

\begin{table}[!htb]
\caption{Input is several sentences from the beginning of a review sample from IMDB Review dataset. Continuation is the generated text output from PLDRv5-DAG-3 LLM model with nucleus sampling at $\text{top-p}=0.8$. The model generates text for 256 tokens or until it encounters an end of sentence ("[END]") token.}
\label{table9}
\centering
\resizebox{\textwidth}{!}{

\begin{tabular}{| p{5cm} | p{5cm} | }
\hline 
\textbf{Input} & \textbf{Continuation} \\ 
\hline
"Cover Girl" is a lacklustre WWII musical with absolutely nothing memorable about it, save for its signature song, "Long Ago and Far Away." This film came out before Gene Kelly really hit his artistic stride, and while there are evidences of his burgeoning talent here, mostly he plays sidekick to Rita Hayworth. And there's the problem. Rita Hayworth is gorgeous, no doubt about that. But she's simply not a compelling screen presence.
&
Here she's hiding her powerful story behind some of the biggest names in Hollywood history and feeling a little letdown. She's got a beautiful one and she's also a series sure to be underwhelming, so it was very good to see her on her debut and hope she comes out on top in the film. The year after her divorce, there's no doubt that Hayworth is a star. I don't know about her, but she has proven to be in the most impressive places in the world and she's earned it for herself. In fact, she's a compelling story that doesn't just happen. I love her performances. The sexual resonance of Hayworth is remarkable. But, as a film star, Hayworth also has a ton of talent. She's not just one actress. She's also a young actress. A production editor for Red Hot Superstars, Hayworth was perhaps one of the best of the best directors on television. But Hayworth will make her name visible on television in a lot of ways. In her first few films, Hayworth has a hard time coming around. And this is where Hayworth is getting at. She has an incredible, bright, thoughtful look and
\\ \hline
\end{tabular} } 
\end{table}

\begin{table}[!htb]
\caption{Input is several sentences from the beginning of a review sample from IMDB Review dataset. Continuation is the generated text output from PLDRv5-DAG-3 LLM model with nucleus sampling at $\text{top-p}=0.8$. The model generates text for 256 tokens or until it encounters an end of sentence ("[END]") token.}
\label{table10}
\centering
\resizebox{\textwidth}{!}{

\begin{tabular}{| p{5cm} | p{5cm} | }
\hline 
\textbf{Input} & \textbf{Continuation} \\ 
\hline
No emotion. Bad music (and I am a reformed eighties metal guy, so I would be vulnerable to some good stuff.) Everything is half done. The brother is a parody, there is nothing to hold us in except to see what horror from the eighties they'll pull up next. The tour manager tries to bring a humanity to the thing but isn't given enough time.
&
The stage is set, the band is packed and it's already about 12-14 days until the concert starts. (If you love dark metal, it's already a groovy force of tenacity.) But the band hasn't stopped going. Just take one look at all the musicians in the band, the last few songs they've played, and they will be back for more than a few more minutes. Their backlists include The Robots, Nirvana, Parachute, Deathtrap, and Mark Your They get their two stage games. As far as the music goes, they're a mess. They're all band members and there are some incredible ways to fill in the gaps and stop this band from delivering on it. If they can't, they can't keep that small group of fans all over the city for a month, in a pitch that has 30 people in front. But in order to stay alive and all that, they have to make an effort to put together the band and how the band can't keep the money. They will do all they can to give back to the community, as long as they keep their schedule with all their friends and family
\\ \hline
\end{tabular} } 
\end{table}

\end{document}